\documentclass[journal]{IEEEtran}
\usepackage{amsmath}
\usepackage{isomath}
\usepackage{color}
\usepackage{graphicx}
\usepackage{subfigure}
\usepackage{amssymb}
\usepackage{siunitx}
\usepackage{booktabs}
\usepackage{tabularx}
\usepackage{comment}
\usepackage{cite}
\usepackage[normalem]{ulem}
\usepackage{threeparttable}

\newtheorem{remark}{Remark}
\newtheorem{lemma}{Lemma}
\newtheorem{theorem}{Theorem}
\newtheorem{corollary}{Corollary}

\newcommand\proof{\textit{Proof:}~}

\newcommand\ti[2]{#1_{\langle{#2}\rangle}}
\newcommand\bti[2]{\boldsymbol{#1}_{\langle{#2}\rangle}}
\newcommand\tli[3]{#1_{#2\langle{#3}\rangle}}
\newcommand\btli[3]{\boldsymbol{#1}_{#2\langle{#3\rangle}}}
\newcommand\tms[3]{\{#1\}_{#2}^{#3}}

\newcommand\ve[1]{\hat{\boldsymbol{#1}}}

\newcommand\bs[1]{\boldsymbol{#1}}

\newcommand\ts[1]{{#1}^{\top}}
\newcommand\tbs[1]{\boldsymbol{#1}^{\top}}
\newcommand\mo[1]{\mathrm{#1}}
\newcommand\eop{$\square$}
\newcommand\figref[1]{Fig. \ref{#1}}

\newcommand\tabref[1]{Table \ref{#1}}
\newcommand\apref[1]{Appendix \ref{#1}}
\renewcommand\eqref[1]{Eq. (\ref{#1})}

\DeclareMathOperator{\diag}{diag}

\definecolor{darkblue}{rgb}{0.8,0,0.0}

\definecolor{backgroundcolor}{RGB}{240, 220, 200}

\makeatletter
\def\endthebibliography{%
  \def\@noitemerr{\@latex@warning{Empty `thebibliography' environment}}%
  \endlist
}
\makeatother

\begin{document}

\title{SRIBO: An Efficient and Resilient Single-Range and Inertia Based Odometry for Flying Robots}

\author{Wei~Dong,~Zheyuan~Mei,~Yuanjiong~Ying,~Sijia~Chen,~Yichen~Xie,~and~Xiangyang~Zhu
\thanks{The authors are with the State Key Laboratory of Mechanical System and Vibration, School of Mechanical Engineering, Shanghai Jiao Tong University, Shanghai 200240, China (e-mail: \{dr.dongwei, chushengbajinban, joyying0222, scarlettchen, slavic, mexyzhu\}@sjtu.edu.cn). Zheyuan Mei and Yuanjiong Ying contributed equally.}
\thanks{Corresponding author: Xiangyang Zhu (mexyzhu@sjtu.edu.cn). }
}

\markboth{Wei Dong et. al. @ CIUS 2022}%
{Shell \MakeLowercase{\textit{et al.}}: Bare Demo of IEEEtran.cls for IEEE Journals}

\maketitle

\begin{abstract}
Positioning with one inertial measurement unit and one ranging sensor is commonly thought to be feasible only when trajectories are in certain patterns ensuring observability. For this reason, to pursue observable patterns, it is required either exciting the trajectory or searching key nodes in a long interval, which is commonly highly nonlinear and may also lack resilience. Therefore, such a positioning approach is still not widely accepted in real-world applications. To address this issue, this work first investigates the dissipative nature of flying robots considering aerial drag effects and re-formulates the corresponding positioning problem, which guarantees observability almost surely. On this basis, a dimension-reduced wriggling estimator is proposed accordingly. This estimator slides the estimation horizon in a stepping manner, and output matrices can be approximately evaluated based on the historical estimation sequence. The computational complexity is then further reduced via a dimension-reduction approach using polynomial fittings. In this way, the states of robots can be estimated via linear programming in a sufficiently long interval, and the degree of observability is thereby further enhanced because an adequate redundancy of measurements is available for each estimation. Subsequently, the estimator's convergence and numerical stability are proven theoretically. Finally, both indoor and outdoor experiments verify that the proposed estimator can achieve decimeter-level precision at hundreds of hertz per second, and it is resilient to sensors' failures.
Hopefully, this study can provide a new practical approach for self-localization as well as relative positioning of cooperative agents with low-cost and lightweight sensors.
\end{abstract}


\IEEEpeerreviewmaketitle

\section{Introduction}
Inertial measurement units (IMUs) and ranging sensors such as ultra-wideband radios (UWBs) are lightweight and low-cost \cite{nguyenNTUVIRALVisualinertialranginglidar2021}. Hence, range-inertial odomotry (RIO) with UWBs and IMUs is practically significant for flying robots \cite{xuDecentralizedVisualInertialUWBFusion2020}. More attractively, these sensors are omnidirectional and insusceptible to illumination conditions. The inconvenience is that multiple pre-installed and calibrated anchors are commonly required \cite{muellerFusingUltrawidebandRange2015, caoAccuratePositionTracking2020,jiaCompositeFilteringUWBbased2022}. This hinders autonomous navigation in unknown environments. To tackle this issue, previous researchers attempt to develop localization algorithms using a single UWB anchor \cite{caoAccuratePositionTracking2020, cossetteRelativePositionEstimation2021, wangSingleBeaconBasedLocalization2016}, which is referred as a single-ranging problem in this work. As such algorithms only need to fuse an IMU with a single UWB ranging measurement, pre-installation and calibration of anchors are no longer required. However, only using the IMU and UWB range measurements at a single time instance is insufficient to determine the position of a robot, and a sliding windows filtering (SWF) associated with observability analyses are necessary \cite{batistaSingleBeaconNavigation2010}. Such analyses are originated in the field of autonomous underwater vehicles \cite{arrichielloObservabilityAnalysisSingle2015, batistaSingleRangeAided2011}, and the results are seamlessly applied for flying robots \cite{caoAccuratePositionTracking2020, cossetteRelativePositionEstimation2021}. Typically, the vehicles or robots are modeled as double-integral systems, and corresponding observability matrices are evaluated either linearly or nonlinearly \cite{arrichielloObservabilityAnalysisSingle2015,indiveriSingleRangeLocalization2016}. These works commonly conclude that the observability is guaranteed only when accumulating actuation in arbitrary two axes is not constant or linearly dependent on each other \cite{arrichielloObservabilityAnalysisSingle2015, cossetteRelativePositionEstimation2021}.

Regarding this conclusion, previous researchers attempt to enhance the observability and positioning performance with two approaches. The first one delicately designs controllers to guarantee linearly independent motions in different axes \cite{caoRelativeDockingFormation2020, nguyenDistanceBasedCooperativeRelative2019,nguyenPersistentlyExcitedAdaptive2020}. The second one performs SWF along with observability evaluation \cite{cossetteRelativePositionEstimation2021, shalabyRelativePositionEstimation2021}. As the first approach requires additional control efforts to guarantee positioning performance, this work focuses on the second one.

Existing SWF approaches are commonly nonlinear, and their computational complexity is hyperlinear \cite{cossetteRelativePositionEstimation2021, dongTrajectoryEstimationFlying2022}. Therefore, the sliding window cannot be too long. Unfortunately, a short sliding window may only include small actuation in all the axes, which causes the observability matrix to be ill-conditioned, not to mention the observability itself is not theoretically guaranteed, as discussed already. To address this issue, Ref. \cite{cossetteRelativePositionEstimation2021} tries to perform a key-node selection approach to ehance the observability in an expanded long sliding window. This approach is practically significant, but there are still two aspects requiring further investigation. First, compared to previous research, there is no difference in the observability condition. Therefore, it has to repeatedly compute the observability matrices and their inverse to find proper key nodes, which could be time-consuming \cite{farrellAidedNavigationGPS2008}. Second, the key-node selection is inherently a desampling and has unavoidable information loss. Therefore, it is still necessary to further pursue a computationally efficient and observability guaranteed estimation approach for real-world application \cite{dongTrajectoryEstimationFlying2022}.

In view of the state-of-the-art, we propose a framework of single-range and inertia based odometry (SRIBO) to enhance the observability with a refined model considering aerodynamics and improve its efficiency in long sling window with a dimension-reduced linear estimator. First, we refine the model of flying robots considering aerial drag effects, and the single-range inertial odometry (SRIO) merely using an IMU and a UWB is therewith formulated and proven to be observable almost surely. In such a case, it is theoretically guaranteed that one can effectively estimate states taking measurements in an arbitrary horizon. Subsequently, a dimension-reduced wriggling estimator (DWE) is proposed. This estimator slides the estimation horizon in a stepping manner, and output matrices are approximately evaluated based on a sequence of historical estimations. On this basis, we can formulate the optimal state estimation as a batched linear least-square programming. Subsequently, a polynomial fitting approach is introduced to reduce the dimension as well as the computational complexity of the estimation. In this manner, the SRIO  can estimate states in a sufficiently long interval with redundant measurements, which further improves the robustness of the estimation. The SRIO is also demonstrated to able to fuse with an additional optical flow sensor, and a so-called single-range, inertia, and optical-flow odometry (SRIFO) is performed. The optical flow sensor provides additional velocity measurements to enhance the estimation performance further, yet failures of such measurements will not lead to singular solutions. Theoretically, both the SRIO and SRIFO in our SRIBO framework are proven to be numerically stable, asymptotic converging and fault-tolerant to sensor failures. Real-time experiments are conducted at last to verify the declarations of observability, computational efficiency, and resilience.

The main contributions are two aspects: 1) modeling with aerial drag effects, the flying robot positioning with SRIBO algorithms is first formulated to be observable almost surely; 2) owing to its efficiency and resilience, the proposed linear DWE can perform estimation in a sufficiently long interval with redundant measurements, which enhances the degree of observability as well as the estimation performance.

\section{Related work}

\subsection{Single-Range and Inertia based Odometry}

The single-range and inertia based odometry has been first considered for the positioning of underwater robots \cite{rossRemarksObservabilitySingle2005,batistaSingleBeaconNavigation2010,ferreiraSingleBeaconNavigation2010}. Practical implementations with Doppler anemometers measuring velocities are also investigated by subsequent researchers \cite{hinsonPathPlanningOptimize2013, wangOptimizationBasedMoving2014}. These positioning approaches are then seamlessly applied for flying robots, e.g., Ref. \cite{nguyenSingleLandmarkDistanceBased2020} uses optical flow sensors measuring velocity alongside a UWB and an IMU. Compared to underwater robots using Doppler anemometers \cite{hinsonPathPlanningOptimize2013, arrichielloObservabilityMetricUnderwater2013, wangOptimizationBasedMoving2014}, the performance of positioning with optical flow sensors may deteriorates with varying illumination or insufficient environmental textures \cite{xuOmniSwarmDecentralizedOmnidirectional2022}. Therefore, an effective positioning approach only using measurements from a UWB and an IMU is still essential to guarantee the robustness of the whole system \cite{cossetteRelativePositionEstimation2021,shalabyRelativePositionEstimation2021}. Besides, such a positioning approach with minimal hardware configuration is practically significant for the relative localization of multiple robots \cite{hanIntegratedRelativeLocalization2019}.

When lacking directly measured velocities, it is promising to introduce derived velocity from the IMU and UWB measurements to improve the positioning performance. For example, in Ref. \cite{caoAccuratePositionTracking2020}, the linear velocity is evaluated to improve positioning precision in a two-dimensional space. Because such an estimation is not precise in more general three-dimensional cases, proper geometric constraints are alternatively imposed in subsequent research regarding the dynamics or kinematics of the robot. For example, position and noise constraints are introduced in Ref. \cite{wangSingleBeaconBasedLocalization2016}. Inherently, position constraints suppress the velocity estimation errors, which further refine the position estimation performance itself \cite{wangSingleBeaconBasedLocalization2016}. To directly constrain the velocity divergence, Ref. \cite{dongTrajectoryEstimationFlying2022} introduces derived constraints to avoid state divergence. Unfortunately, the effectiveness of the derived velocity constraints has not been investigated mathematically, and it should be studied with observability analysis.

\subsection{Observability based Estimation}

In accumulated studies, observability analyses have been either carried out with a classical linearization procedure \cite{batistaSingleBeaconNavigation2010, batistaSingleRangeAided2011, indiveriSingleRangeLocalization2016}, or derived from modern control theories \cite{rossRemarksObservabilitySingle2005, arrichielloObservabilityAnalysisSingle2015,berkaneNonlinearNavigationObserver2021}. Such research commonly concludes that the observability can only be guaranteed when the accumulated excitation is linearly independent on different axes \cite{caoAccuratePositionTracking2020}. Regarding this conclusion, an active control approach with exponential convergence is proposed in Ref. \cite{nguyenDistanceBasedCooperativeRelative2019}, which persistently excites trajectories to guarantee the observability. Similarly, an integrated estimation-control scheme is proposed to achieve asymptotic convergence in the formation control \cite{caoRelativeDockingFormation2020}. Although the aforementioned active control strategies demonstrate their effectiveness, extra control efforts apart from their primary control objectives are commonly required to guarantee the observability.

As an alternative, recent studies have attempted to iteratively select key nodes, which ensures observability in a long sliding window, to improve the state estimation performance \cite{cossetteRelativePositionEstimation2021}. This is intrinsically a state augmentation approach \cite{ferreiraSingleBeaconNavigation2010}, which introduces auxiliary state variables deduced by a combination of concurrent states and/or historical states to enhance the degree of observability \cite{shenQuantifyingObservabilityAnalysis2018}. However, this sliding window filtering approach is highly nonlinear and searching observability-guaranteed key-nodes is also time-consuming \cite{farrellAidedNavigationGPS2008}.

Notably, employing velocity measurements can be also viewed to enhance the observability with extra hardware. As discussed previously, this approach has been verified by underwater robots equipped with Doppler anemometers \cite{hinsonPathPlanningOptimize2013, wangOptimizationBasedMoving2014} and flying robots equipped with optical flow sensors \cite{nguyenSingleLandmarkDistanceBased2020}. In the cooperative positioning of flying robots, the UWB and IMU based state estimation is also enhanced with velocity measurements from optical flow sensors \cite{guoUltraWidebandOdometryBasedCooperative2020,guoUltrawidebandBasedCooperative2017}. In a similar manner, visual odometry is included in Refs. \cite{nguyenTightlycoupledUltrawidebandaidedMonocular2020,nguyenFlexibleResourceEfficientMultiRobot2022, nguyenRangeFocusedFusionCameraIMUUWB2021,zhengUWBVIOFusionAccurate2022a}. More comprehensively, an omnidirectional visual–inertial–UWB framework is further proposed for aerial swarm \cite{xuOmniSwarmDecentralizedOmnidirectional2022}. Compared to these implementations with optical sensors, the SRIO has a minimal hardware configuration and is still worth furhter investigation.

\subsection{Sliding Window Filtering}
The SWF, also named as the moving horizon estimation (MHE) \cite{raoConstrainedLinearState2001, wynnConvergenceGuaranteesMoving2014}, is an optimal state estimation approach using a sliding window of measurements in each time step \cite{dong-siMotionTrackingFixedlag2011}. Its effectiveness and robustness have been well investigated \cite{raoConstrainedStateEstimation2003, wynnConvergenceGuaranteesMoving2014}, and it is also experimentally verified that its performance is better than that of the classical Kalman filter \cite{haseltineCriticalEvalationExtended2005}.
For this reason, the single-range based positioning system is commonly developed in an SWF manner \cite{cossetteRelativePositionEstimation2021, shalabyRelativePositionEstimation2021}. Unfortunately, because of the nonlinearity of the optimization, the computational complexity increases hyperlinearly with the size of the sliding window, which may correlate with the degree of observability. Although the key-node selection approach can reduce the states required to estimate in a long sliding window \cite{cossetteRelativePositionEstimation2021}, it has to iteratively estimate the observability condition, and the computational cost is still considerable. To tackle this issue, a gradient-aware approach to enhance the computational efficiency is proposed in Ref. \cite{dongTrajectoryEstimationFlying2022}. The divergence is also avoided with velocity penalty. However, this approach is not well formulated in a mathematical manner, and it still cannot well handle a long window of estimation. 

When extra sensors, e.g., camera, visual-inertial odometry (VIO), etc, are included alongside the UWB \cite{xuDecentralizedVisualInertialUWBFusion2020,nguyenFlexibleResourceEfficientMultiRobot2022}, as the observability is commonly not considered as a main issue ignoring sensor failures, the sliding window size and the computational efficiency have been not well considered in existing work \cite{zhangAgileFormationControl2022a, nguyenRangeFocusedFusionCameraIMUUWB2021, zieglerDistributedFormationEstimation2021}.

\section{Preliminaries}

\subsection{Notations}
 In this work, $\boldsymbol{0}_{m\times r}$ denotes a zero matrix with dimension $m\times r$. For simplicity, $\boldsymbol{0}_{n}$ denotes a square matrix $\boldsymbol{0}_{n\times n}$. Similarly, $\boldsymbol{I}_{n}$ denotes an identity matrix with dimension $n\times n$. We may further omit the subscript when there is no ambiguity. For a matrix $\boldsymbol{A}$, $\boldsymbol{A}_{ij}$ denotes the element in the $i$th row and $j$th column. $\boldsymbol{A}_i$ denotes the vector from $i$th column of $\boldsymbol{A}$. $\boldsymbol{A}_{[i_1:i_2,j_1:j_2]}$ denotes a block matrix with elements extracted from $i_1$th to $i_2$th rows and $j_1$th to $j_2$th columns in matrix $\bs{A}$.  A diagonal matrix is denoted as $\boldsymbol{D}=\operatorname{diag}(\boldsymbol{d})$, with all elements zeros except $\boldsymbol{D}_{ii}=\boldsymbol{d}_i$.  The integer number set is denoted as $\mathbb{Z}$, and the natural number set is denoted as $\mathbb{N}$, which is equivalent to the positive integer number set $\mathbb{Z}_+$. A number $x$ is said to belong to $\mathbb{N}_{>N}$ when $x,N\in\mathbb{N}$ and $x>N$. Similarly, $x\in\mathbb{N}_{\geq N}$ when $x, N\in\mathbb{N}$, and $x\geq N$. A number $x$ is said to belong to $\mathbb{N}_{[N_1,N_2]}$ when $x, N_1, N_2\in \mathbb{N}$ and $ N_1\leq x\leq N_2$.

 The flying robot is supposed to fly in a $d$-dimensional space ($d\in \mathbb{N}_{>1}$). We use $\hat{\bs{x}}$ and $\bar{\bs{x}}$ to denote the estimation and measurement, respectively,  of any state variable $\bs{x}$ of the robot. we further use $\boldsymbol{x}_i$ to index the $i$th element of $\boldsymbol{x}$. Meanwhile, $\ti{\boldsymbol{x}}{i}$ is adopted to denote the value of $\boldsymbol{x}$ at some time instant $i$ or $t_i$. Combinationally, $\btli{{x}}{j}{i}$ represents the value of $\boldsymbol{x}_j$ at instant $i$ or $t_i$. A sequence of states continuously sampled at a certain time interval is abbreviated as $\bti{x}{[{k_1:k_2}]}\triangleq \{{\bs{x}}\}_{k_1}^{k_2}\triangleq \{\ti{\bs{x}}{k_1},\ti{\bs{x}}{k_1+1}, \cdots, \ti{\bs{x}}{k_2}\}$.

\subsection{Multirotor Flying Robots with Aerial Drags}

In this section, we first investigate the observability of a multirotor flying robot considering aerial drag effects. As investigated by Refs. \cite{martinTrueRoleAccelerometer2010a,leishmanQuadrotorsAccelerometersState2014,svachaImprovingQuadrotorTrajectory2017a}, a linear drag force equation is adequate to formulate such effects. That is, $\boldsymbol{f}_{D}=-m\boldsymbol{\mu} \boldsymbol{v}$, where $\boldsymbol{f}_{D}$ is the drag force, $m$ is the mass of the robot, $\boldsymbol{v}$ is the velocity vector, and $\boldsymbol{\mu}$ is a diagonal matrix representing the drag coefficients in different directions.

Taking the ranging odometry as the output, the state space representation of the SRIO problem is written as follows
\begin{equation}
 \left\{\begin{array}{l}\dot{\boldsymbol{x}}=\left[\begin{array}{ll}\mathbf{0}_{d} & \boldsymbol{I}_{d} \\
 \mathbf{0}_{d} & -\boldsymbol{\mu}\end{array}\right] \boldsymbol{x}+\left[\begin{array}{c}\mathbf{0}_{d} \\\boldsymbol{I}_{d}\end{array}\right] \boldsymbol{u} + \bs{n_x} \\ y=h(\boldsymbol{x})=\frac{1}{2}\|\boldsymbol{p}\|^{2}+{n_y}\end{array}\right.
 \label{eq:modelwithdrag}
\end{equation}
where $\bs{n_x}$ is the process noise, ${n_y}$ is the measurement noise, $\boldsymbol{p}$ is the position of the robot, $\boldsymbol{x}=[\ts{\boldsymbol{p}},\ts{\boldsymbol{v}}]^{\top}$, and $\bs{u}$ is a $d\times 1$ input vector which physically corresponds to the acceleration of the flying robots.

If an additional optical flow sensor is available, one can further introduce the velocity of the robot into the measurement and implement an SRIFO system, and the state space representation is re-formulated as
\begin{equation}
 \left\{\begin{array}{l}\dot{\boldsymbol{x}}=\left[\begin{array}{ll}\mathbf{0}_{d} & \boldsymbol{I}_{d} \\
 \mathbf{0}_{d} & -\boldsymbol{\mu}\end{array}\right] \boldsymbol{x}+\left[\begin{array}{c}\mathbf{0}_{d} \\\boldsymbol{I}_{d}\end{array}\right] \boldsymbol{u} + \bs{n_x} \\ \bs{y}=\bs{h}(\boldsymbol{x})=[\frac{1}{2}\|\boldsymbol{p}\|^{2}, \ts{\bs{v}}]^{\top}+\bs{n_y}\end{array}\right.
 \label{eq:dragmodelwithflow}
\end{equation}

Either \eqref{eq:modelwithdrag} or \eqref{eq:dragmodelwithflow} can be abstractly described with the following compact form
\begin{equation}
\dot{\boldsymbol{x}}=\boldsymbol{f}(\boldsymbol{x}, \boldsymbol{u})+ \bs{n_x} ,\ \bs{y}=\bs{h}(\boldsymbol{x})+ \bs{n_y}
\label{eq:absmodelnonlinear}
\end{equation}

\begin{remark}
In practice, a state estimator is commonly implemented independently from the controller. For the sake of convenience, the control input variable $\bs{u}$ can be evaluated from the measurement of the IMU.
\end{remark}

The aforementioned SRIBO systems are nonlinear, they are locally weakly observable only if the observability matrix
\begin{equation}
\mathcal{O}^{\top}:=[\nabla \mathcal{L}_{\bs{f}}^{0} \bs{h}, \nabla \mathcal{L}_{\bs{f}}^{1} \bs{h},  \cdots, \nabla \mathcal{L}_{\bs{f}}^{k} \bs{h}]
\label{eq:localobscdn}
\end{equation}
is full rank \cite{hermannNonlinearControllabilityObservability1977a}. Here $\nabla$ denotes the gradient operator and $\mathcal{L}_{\bs{f}}^{\alpha} \bs{h}$ is the set of the $\alpha$-order Lie derivatives. Specifically, $\mathcal{L}_{\bs{f}}^{\alpha} \bs{h} = \nabla[\mathcal{L}_{\bs{f}}^{\alpha-1} \bs{h}]\cdot \bs{f}$, and $\mathcal{L}_{\bs{f}}^{0} \bs{h} = \bs{h}$.

\section{Dimension-reduced Wriggling Estimator}

In this section, the DWE is proposed to efficiently estimate states of SRIBO systems. We first derive this approach considering the SRIO problem and then extend it to the SRIFO problem. Its numeric stability, convergence, and fault-tolerance properties are also theoretically investigated.

\subsection{The DWE for the SRIO}

\label{sec:srio}

\subsubsection{Observability Analysis} When formulated with the aerial drag effects, the observability of an SRIO system is guaranteed by the following theorem.

\begin{theorem}
\label{th:obsiro}
 The SRIO system described with \eqref{eq:modelwithdrag} is locally weakly observable almost surely.
\end{theorem}

\proof According to the condition stated in \eqref{eq:localobscdn}, the observability matrix of \eqref{eq:modelwithdrag} is written as
\begin{equation}
\small
\mathcal{O}=\left[\begin{array}{cc}\ts{\bs{p}} & \boldsymbol{0}_{1\times d}    \\
 \ts{\bs{v}} & \ts{\bs{p}} \\
 (\boldsymbol{u}-\boldsymbol{\mu} \boldsymbol{v})^{\top}  & (2 \boldsymbol{v}-\boldsymbol{\mu}\boldsymbol{p})^{\top}\\
-\boldsymbol{\mu}(\boldsymbol{u}-\boldsymbol{\mu} \boldsymbol{v})^{\top} & (3 \boldsymbol{u}-6\boldsymbol{\mu} \boldsymbol{v}+\boldsymbol{\mu}^{2}\boldsymbol{p})^{\top}\\
 \boldsymbol{\mu}^{2}(\boldsymbol{u}-\boldsymbol{\mu} \boldsymbol{v})^{\top}
    & (-10 \boldsymbol{\mu}\boldsymbol{u}+14\boldsymbol{\mu}^{2} \boldsymbol{v}-\boldsymbol{\mu}^{3}\boldsymbol{p})^{\top} \\
    - \boldsymbol{\mu}^{3}(\boldsymbol{u}-\boldsymbol{\mu} \boldsymbol{v})^{\top}
    & (25 \boldsymbol{\mu}^{2}\boldsymbol{u}-30\boldsymbol{\mu}^{3} \boldsymbol{v}+\boldsymbol{\mu}^{4}\boldsymbol{p})^{\top} \\ \end{array}\right]
    \label{eq:obsrio}
\end{equation}
the detailed derivation of which is presented in \apref{ap:lie_srio}.

When denoting $\bs{\chi_{\mu}} = [{\bs{p}},{\bs{v}},{\bs{u}}-\bs{\mu}\bs{v}]^{\top}$, one can have $ \mathrm{rank}(\mathcal{O})< 2d \iff \bs{\chi_{\mu}}_i \equiv \alpha_j \bs{\chi_{\mu}}_j+\alpha_k \bs{\chi_{\mu}}_k, \bs{\mu}_i \equiv \bs{\mu}_j \equiv \bs{\mu}_k$ ($i, j, k\in[1,d]$). This means only when the position, velocity, and acceleration in one direction is always a linear combination of that in the other two directions, this SRIO system is unobservable. This is hardly seen in real-world autonomous navigation. Therefore, probabilistically $p(\mathrm{rank}(\mathcal{O})= 2d)=1$. That is, almost surely the observability matrix in \eqref{eq:obsrio} is full rank, and accordingly the SRIO system is locally weakly observable almost surely.
\eop

\begin{remark}
 It should be mentioned that an absolutely static state $\bs{v}\equiv\bs{u}\equiv\bs{0}$ is also a special unobservable case (viewing the condition $\bs{\chi_{\mu}}_i \equiv \alpha_j \bs{\chi_{\mu}}_j+\alpha_k \bs{\chi_{\mu}}_k, \bs{\mu}_i \equiv \bs{\mu}_j \equiv \bs{\mu}_k$). Although we cannot find an absolutely static flying robot in real-flights, but quasi-static states are possibly observed. Apparently, when flying robots in quasi-static state, the condition number of the observability matrix significantly increases. Therefore, the estimation performance will highly correlate with flight velocities and accelerations near quasi-static scenarios.
\end{remark}

\begin{remark}
 If the aerial drag force is not included, i.e., $\bs{\mu}\equiv \bs {0}$, the observability matrix is formulated as
\begin{equation}
 \ts{\mathcal{O}}=\left[\begin{array}{cccccc}\bs{p} &  \bs{v} & \bs{u} & \boldsymbol{0}_{n \times 1} &  \boldsymbol{0}_{n \times 1}& \boldsymbol{0}_{n \times 1} \\  \boldsymbol{0}_{n \times 1} & \bs{p} &  2\bs{v} & 3\bs{u} &  \boldsymbol{0}_{n \times 1}&  \boldsymbol{0}_{n \times 1} \\ \end{array}\right]
 \end{equation}
 which is rank deficiency, i.e., $\mo{rank}({\mathcal{O}})\leq 4$. This is why accumulating works commonly conclude that the SRIO system is not locally weakly observable \cite{arrichielloObservabilityAnalysisSingle2015}.
\end{remark}

 Although the SRIO system is observable almost surely, the degree of observability may vary significantly in real-time applications. In particular, when the motion pattern of one direction is closely linearly dependent on that of the other directions, the observability matrix $\mathcal{O}$ is ill-conditioned, which will further lead to estimation performance deterioration. For this reason, we further develop the DWE ensuring numeric stability.

\subsubsection{DWE}\label{sec:dwe} For a convenient implementation, we re-write the the observation model in \eqref{eq:modelwithdrag} as $y=\|\bs{p}\|+n_y$. Subsequently, neglecting noise terms, a discrete model can be formulated by linearizing with a sampling period $\mo{d}t$
\begin{equation}
 \left\{\begin{array}{l}\ti{\boldsymbol{x}}{k+1}=\bti{A}{k} \ti{\boldsymbol{x}}{k}+ \bti{B}{k} \ti{\boldsymbol{u}}{k} \\ \ti{y}{k+1}=\bti{C}{k+1}\ti{\boldsymbol{x}}{k+1}\end{array}\right.
 \label{eq:discretemodel}
\end{equation}
with $\ti{\bs{A}}{k}=\left[\begin{array}{ll}\bs{I}_{d} & \mo{d}t\boldsymbol{I}_{d} \\
 \mathbf{0}_{d} & \bs{I}_{d}-\mo{d}t\boldsymbol{\mu}\end{array}\right]$, $\ti{\bs{B}}{k}=\left[\begin{array}{c}\frac{1}{2}\mo{d}t^2 \\ \mo{d}t \\ \end{array}\right]\otimes\bs{I}_{d}$, and $\bti{C}{k}=[\ti{\bs{p}^{\top}}{k}/\|\bti{p}{k}\|, \bs{0}_{1\times d}]$. Because both $\ti{\bs{A}}{k}$ and $\ti{\bs{B}}{k}$ are time invariant for some fixed $\mo{d}t$, we use abbreviations $\bs{A}\triangleq \ti{\bs{A}}{k}$ and $\bs{B}\triangleq \ti{\bs{B}}{k}$ in the following formulation when without confusion.

\begin{figure}
  \centering
  \includegraphics[width=0.95\columnwidth]{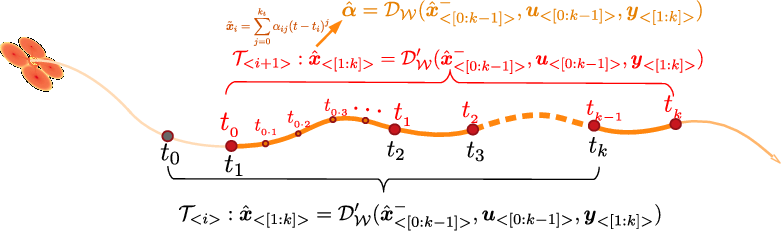}
  \caption{A conceptual illustration of dimension-reduced wriggling estimator}
  \label{fig:slidingscheme}
\end{figure}

 On this basis, the DWE $\mathcal{D_W}$ is implemented and conceptually illustrated in Fig. \ref{fig:slidingscheme}. Upon each period $\ti{\mathcal{T}}{i+1}$, we aim to perform {\it{a maximum a posteriori}} estimate of the state sequence $\{\hat{\bs{x}}\}_{1}^k$ based on the input $\{\bs{u}\}_0^{k-1}$, the range measurement $\tms{\bar{r}}{1}{k}$, and a prior estimate $\tms{\hat{\bs{x}}^{-}}{0}{k-1}$ which is equivalent to the posteriori estimation $\tms{\hat{\bs{x}}}{1}{k}$ in previous period $\ti{\mathcal{T}}{i}$. Mathematically, this estimator is implemented as follows in an interval with $k_w+1\ (k_w\in\mathbb{N}_{>2d})$ sampling points
\begin{equation}
 \hat{\boldsymbol{x}}=\arg \min_{\boldsymbol{x}} J= \sum_{i=0}^{k_w-1}\tli{c}{1}{i} + \sum_{i=1}^{k_w}\tli{c}{2}{i} + \sum_{i=1}^{k_w}\tli{c}{3}{i}
 \label{eq:generaslidingproblem}
\end{equation}
where $\tli{c}{1}{i}=\btli{e}{1}{i}^{\top}\ti{\bs{P}^{-1}}{{i}}\tli{\bs{e}}{1}{i}$, $\tli{c}{2}{i}=\btli{e}{2}{i}^{\top}\ti{\bs{Q}^{-1}}{{i}}\tli{\bs{e}}{2}{i}$, $\tli{c}{3}{i}=\btli{e}{3}{i}^{\top}\ti{\bs{R}^{-1}}{{i}}\tli{\bs{e}}{3}{i}$, $\btli{e}{1}{i}\triangleq \ti{\ve{x}}{i}-\ti{\ve{x}}{i}^{-}$, $\btli{e}{2}{i} \triangleq \ti{\ve{x}}{i}-\boldsymbol{A}\ti{\ve{x}}{i-1}-\boldsymbol{B}{\bti{u}{i-1}}$, and $\btli{e}{3}{i}\triangleq \ti{\bar{r}}{i}-\bti{C}{i}\ti{\ve{x}}{i}$. Statistically, $\bti{P}{i}$, $\bti{Q}{i}$ and $\bti{R}{i}$ correspond to the covariance matrices of the state estimate, the process, and the measurement, respectively \cite{dongTrajectoryEstimationFlying2022}.

\begin{remark}
For typical flying robots such as multirotor robots, the input $\bti{u}{i}$ equals the instant acceleration $\bti{a}{i}$, which can be directly measured by the IMU. The lateral acceleration $\bs{a}_x$ and $\bs{a}_y$ are also closely related to the attitude of the flying robot when the motion is not too aggressive. Specifically, $\bs{a}_x\approx g\sin \theta\approx g\theta$, and $\bs{a}_y\approx -g\sin \phi \approx -g \phi $, where $\theta$ and $\phi$ are the pitch and roll angles \cite{dongHighperformanceTrajectoryTracking2014}. In practice, the attitude estimation precision is higher than that of the directly measured acceleration. Therefore, it is a good choice to evaluate the acceleration based on the attitude estimation.
\end{remark}

Denoting $\boldsymbol{E}=[\tli{\boldsymbol{e}^{\top}}{1}{0}, \tli{\boldsymbol{e}^{\top}}{1}{1}, \cdots, \tli{\boldsymbol{e}^{\top}}{1}{k_w-1}, \tli{\boldsymbol{e}^{\top}}{2}{1}, \tli{\boldsymbol{e}^{\top}}{2}{2}, \cdots$, $\tli{\boldsymbol{e}^{\top}}{2}{k_w}$, $\tli{\boldsymbol{e}^{\top}}{3}{1}, \tli{\boldsymbol{e}^{\top}}{3}{2}, \cdots, \tli{\boldsymbol{e}^{\top}}{3}{k_w}]$, $\boldsymbol{P}=\diag (\bti{P}{0}, \bti{P}{1}$, $\cdots$, $\bti{P}{k_w-1})$, $\boldsymbol{Q}=\diag (\bti{Q}{0}, \bti{Q}{1}, \cdots, \bti{Q}{k_w-1})$, $\boldsymbol{R}=\diag (\bti{R}{1}, \bti{R}{2}, \cdots, \bti{R}{k_w})$, and $\boldsymbol{W}=\diag (\boldsymbol{P}^{-1}, \boldsymbol{Q}^{-1}, \boldsymbol{R}^{-1})$, \eqref{eq:generaslidingproblem} can be re-formulated as a batch nonlinear least-squares estimator as follows \cite{barfootStateEstimationRobotics2017}
\begin{equation}
  \hat{\boldsymbol{x}}=\arg \min_{\boldsymbol{x}} J = \boldsymbol{E}^{\top}\boldsymbol{W}\boldsymbol{E}
  \label{eq:batchls}
\end{equation}

The problem formulated with \eqref{eq:generaslidingproblem} or \eqref{eq:batchls} is general solved with the Gauss-Newton algorithm or its variants \cite{dongTrajectoryEstimationFlying2022,cossetteRelativePositionEstimation2021}. It is time-consuming when the window size, i.e., $k_w$, becomes large. Therefore, this work develops an alternative to solve this problem with high efficiency.

To facilitate this development, based on \eqref{eq:discretemodel} and \eqref{eq:generaslidingproblem}, the matrix $\bs{E}$ is re-written as
\begin{equation}
 \boldsymbol{E}=
 \begin{bmatrix}
\tilde{\boldsymbol{I}}_A \\
\tilde{\boldsymbol{A}}\\
\tilde{\boldsymbol{C}}\\
 \end{bmatrix}
 \boldsymbol{x}-
  \begin{bmatrix}
\boldsymbol{I}_{2dk_w} & \boldsymbol{0} & \boldsymbol{0} \\
\boldsymbol{0} & \tilde{\boldsymbol{B}} & \boldsymbol{0}\\
\boldsymbol{0} & \boldsymbol{0} & \boldsymbol{I}_{k_w}\\
 \end{bmatrix}\boldsymbol{\theta}
 \triangleq
 \bs{E_x}{\bs{x}}-\bs{E_{\theta}}\bs{\theta}
 \label{eq:costmatrix}
\end{equation}
where $\boldsymbol{x}=[\ti{\bs{x}^{\top}}{0}, \ti{\bs{x}^{\top}}{1}, \cdots, \ti{\bs{x}^{\top}}{k_w}]$, $\boldsymbol{\theta}=[\ti{\hat{\bs{x}}^{-\top}}{0}, \ti{\hat{\bs{x}}^{-\top}}{1}$, $\cdots$, $\ti{\hat{\bs{x}}^{-\top}}{k_w-1}, \ti{\bs{a}^{\top}}{0}, \ti{\bs{a}^{\top}}{1},\cdots, \ti{\bs{a}^{\top}}{k_w-1}, \ti{\bar{r}}{1}, \ti{\bar{r}}{2}, \cdots, \ti{\bar{r}}{k_w}]^{\top}$, $\tilde{\boldsymbol{I}}_A = [\boldsymbol{I}_{2dk_w},\ \boldsymbol{0}_{2dk_w\times 2d}]$, $\tilde{\boldsymbol{A}}$ is a block matrix with $\tilde{\boldsymbol{A}}_{[2di+1:2d(i+1),2di+1:2d(i+2)]}=[-\bs{A},\ \bs{I}_{2d}]\ (i\in [0, k_w-1])$, $\tilde{\boldsymbol{C}}$ is a block matrix $\tilde{\boldsymbol{C}}_{[i,2di+1:2d(i+1)]} = \bti{C}{i}\ (i\in[1,k_w])$, and $\tilde{\boldsymbol{B}}$ is a diagonal matrix with $\tilde{\boldsymbol{B}} = \mo{diag}([\underbrace{\bs{B}, \bs{B}, \cdots, \bs{B}}_{k_w\times \bs{B}}])$.

Reasonably, $\bti{C}{i}=\bti{C}{i}^{-},\ i \in [1, k_w-1]^{\top}$, where $\bti{C}{i}^{-}$ equals $\bti{C}{i+1}$ in previous period. Particularly, $\bti{C}{k_w}$ can be approximated with $\bti{C}{k_w-1}$, or we can first evaluate a prior estimate $\tilde{\bs{x}}_{k_w} = \bs{A}\ve{x}_{k_w-1}+\bs{B}\bs{u}$, and then evaluate $\bti{C}{k_w}$ with \eqref{eq:discretemodel}. On this basis, substituting \eqref{eq:costmatrix} into \eqref{eq:batchls}, one can obtain an optimal estimation $\hat{\bs{x}}$ by taking $\frac{\partial{J}}{\partial\bs{x}}=0$. An analytical solution to this problem is
\begin{equation}
\hat{\bs{x}}=(\bs{E_x}^{\top}\bs{W}\bs{E_x})^{-1}\bs{E_x}^{\top}\bs{W}\bs{E_{\theta}}\bs{\theta}
\label{eq:siro_direct}
\end{equation}

\begin{lemma}
  Denoting $\bs{W_E}\triangleq\bs{E_x}^{\top}\bs{W}\bs{E_x}$, if $\bti{P}{i}$ and $\bti{Q}{i}$ are positive definite, the matrix $\bs{W_E}$ is always invertible and the solution of \eqref{eq:siro_direct} exists uniquely.
  \label{lm:invertwe}
\end{lemma}

\proof According to the definition, we have
\begin{equation}
\boldsymbol{W_E}=\tilde{\boldsymbol{I}}_A^{\top} \boldsymbol{P}^{-1} \tilde{\boldsymbol{I}}_A+\tilde{\boldsymbol{C}}^{\top} \boldsymbol{R}^{-1} \tilde{\boldsymbol{C}}+\tilde{\boldsymbol{A}}^{\top} \boldsymbol{Q}^{-1} \tilde{\boldsymbol{A}}
\end{equation}

Apparently, for an arbitrary nonzero vector $\bs{x}$, $\boldsymbol{x}^{\top}\boldsymbol{W_E}\boldsymbol{x}\geq 0$.
The equality establishes if and only if $\boldsymbol{x}^{\top}\tilde{\boldsymbol{A}}^{\top} \boldsymbol{Q}^{-1} \tilde{\boldsymbol{A}}\boldsymbol{x}=\boldsymbol{x}^{\top}\tilde{\boldsymbol{I}}_A^{\top} \boldsymbol{P}^{-1} \tilde{\boldsymbol{I}}_A\boldsymbol{x}=\boldsymbol{x}^{\top}\tilde{\boldsymbol{C}}^{\top} \boldsymbol{R}^{-1} \tilde{\boldsymbol{C}} \boldsymbol{x}=\bs{0}$. As the null space of $\tilde{\boldsymbol{A}}^{\top} \boldsymbol{Q}^{-1} \tilde{\boldsymbol{A}}$ is spanned by $\bs{N_A}\boldsymbol{x}_Q\triangleq[\bs{I},\boldsymbol{A}^{\top},...,\boldsymbol{A}^{k\top}]^{\top}\boldsymbol{x}_Q$ , and the null space of $\tilde{\boldsymbol{I}}_A^{\top} \boldsymbol{P}^{-1} \tilde{\boldsymbol{I}}_A$ is $\mo{span}(\boldsymbol{0}^{\top},\boldsymbol{0}^{\top},...,\boldsymbol{0}^{\top},\boldsymbol{x}_P^{\top})^{\top}$, where $\boldsymbol{x}_Q\neq\bs{0}$ and $\boldsymbol{x}_P\neq\bs{0}$, one can have
\begin{equation}
\mo{Null}(\tilde{\boldsymbol{A}}^{\top} \boldsymbol{Q}^{-1} \tilde{\boldsymbol{A}})\cap \mo{Null}(\tilde{\boldsymbol{I}}_A^{\top} \boldsymbol{P}^{-1} \tilde{\boldsymbol{I}}_A)= \emptyset
\end{equation}

Therefore, $\boldsymbol{W_E}$ is positive definite. On this basis, $\boldsymbol{W_E}$ is invertible, and the solution of \eqref{eq:siro_direct} exists uniquely.
\eop

As stated in Lemma \ref{lm:invertwe}, the invertibility of $\bs{W_E}$ does not depend on the observability of the original system (\eqref{eq:discretemodel}). This is because we re-use a long window of estimated states in the previous period, and their weights contribute a positive semi-definite matrix $\bs{P}^{-1}$.
If we don't re-use those estimated states, then the invertibility of $\bs{W_E}$ depends on the observability of the original system. This is stated as the following corollary.

\begin{corollary}
  \label{corol:weinvert}
  When \eqref{eq:discretemodel} is observable, and $k_w \in \mathbb{N}_{\geq 2d}$, $\bs{W_E}$ is invertible and its invertibility is independent from the choice of $\bs{P}$.
\end{corollary}

\proof Selecting an arbitrary vector $\boldsymbol{x}=[\boldsymbol{x}_Q^{\top},\boldsymbol{x}_Q^{\top}\boldsymbol{A}^{\top},...,\boldsymbol{x}_Q^{\top}(\boldsymbol{A}^k)^{\top}]^{\top}$ from $\mo{Null}(\tilde{\boldsymbol{A}}^{\top} \boldsymbol{Q}^{-1} \tilde{\boldsymbol{A}})$, and substituting it into $\boldsymbol{x}^{\top}\tilde{\boldsymbol{C}}^{\top} \boldsymbol{R}^{-1} \tilde{\boldsymbol{C}} \boldsymbol{x}$, one can have
\begin{equation}
\boldsymbol{x}^{\top}\tilde{\boldsymbol{C}}^{\top} \boldsymbol{R}^{-1} \tilde{\boldsymbol{C}} \boldsymbol{x}
=\sum_{i=0}^{k}\boldsymbol{G}_i^{\top} \boldsymbol{R}_i^{-1}\boldsymbol{G}_i\geq0
\end{equation}
where $\boldsymbol{G}_i=\boldsymbol{C}_i\boldsymbol{A}^{i}\boldsymbol{x}_Q$.

When the system described by \eqref{eq:discretemodel} is observable, one can also have a full rank observability matrix regarding this discrete model, i.e.,
$\mo{rank}(\mathcal{O}_d)= 2d,\ (\mathcal{O}_d=\left[\bti{C}{1}^{\top},
\tbs{A}\bti{C}{2}^{\top}, \cdots, \boldsymbol{A}^{2d-1\top}\bti{C}{2d}^{\top}\right]^{\top})$. On this basis, $\mathcal{O}_d\bs{A}\bs{x}_Q\neq \bs{0}$. Therefore, when $k_w \in \mathbb{N}_{\geq 2d}$, $\tilde{\boldsymbol{C}}^{\top} \boldsymbol{R}^{-1} \tilde{\boldsymbol{C}}$ is positive definite and is invertible. Equivalently, when $k_w \in \mathbb{N}_{\geq 2d}$, $\bs{W_E}$ is invertible and its invertibility is independent from the choice of $\bs{P}$.
\eop

Lemma \ref{lm:invertwe} ensures the numeric stability of the SRIO. That is, even if the original system is at an unobservable point (recall that the original system is observable almost surely), the solution of the SRIO always exists. In contrast, corollary \ref{corol:weinvert} implies that even without pre-estimated states $\bs{x}^{-}$, we can start from an arbitrary point that is observable and obtain a quasi-optimal state estimation with local measurements.

The convergence of the SRIO is further guaranteed by the following theorem.

\begin{theorem}
  \label{th:siro_org}
  The estimator proposed in \eqref{eq:siro_direct} is asymptotically stable almost surely.
\end{theorem}

\proof Regarding \eqref{eq:costmatrix}, we denote $\bs{W_{\theta}}\triangleq\bs{E_x}^{\top}\bs{W}\bs{E_{\theta}}$, which can be explicitly formulated as $\bs{{W}_{\theta}} = [ \tilde{\boldsymbol{I}}_A^{\top} \bs{P}^{-1},\  \tilde{\boldsymbol{A}}^{\top}\bs{Q}^{-1}\tilde{\boldsymbol{B}},\ \tilde{\boldsymbol{C}}^{\top}\bs{R}^{-1}]$.

If a deviation $\delta \ve{x}$ exists in the estimation, we can obtain the following error propagating relationship
\begin{equation}
\delta\hat{\bs{x}}=\bs{W_E}^{-1}(\tilde{\boldsymbol{I}}_A^{\top} \bs{P}^{-1}\tilde{\boldsymbol{I}}_A)\tilde{\boldsymbol{I}}_A^{\top}\delta\hat{\bs{x}}^{-}
\end{equation}

In this work, we select $\bs{P}$, $\bs{Q}$ and $\bs{R}$ empirically, and $\bs{A}$ is a constant matrix. Therefore, they are not affected by the estimation deviation $\delta \hat{\bs{x}}$. In contrast, the output matrix $\tilde{\bs{C}}$ will deviate from its ground truth due to the deviation $\delta \hat{\bs{x}}$. Nonetheless, according to the conclusion of Theorem \ref{th:obsiro} and Corollary \ref{corol:weinvert}, with an arbitrary output matrix $\tilde{\bs{C}}$, $\tilde{\boldsymbol{A}}^{\top}\bs{Q}^{-1}\tilde{\boldsymbol{A}}+\tilde{\boldsymbol{C}}^{\top}\bs{R}^{-1}\tilde{\boldsymbol{C}}$ is positive definite almost surely.

In such a case, as $\tilde{\boldsymbol{I}}_A^{\top} \bs{P}^{-1}\tilde{\boldsymbol{I}}_A$ is semi-positive definite, the spectral radius $\rho\left((\tilde{\boldsymbol{I}}_A^{\top} \bs{P}^{-1}\tilde{\boldsymbol{I}}_A)^{-1}\bs{W_E}\right)>1$, or equivalently, $\rho(\bs{W_E}^{-1}\tilde{\boldsymbol{I}}_A^{\top} \bs{P}^{-1}\tilde{\boldsymbol{I}}_A)< 1$ almost surely. Hence, the estimation error $\delta\hat{\bs{x}}$ will asymptotically converge to zero, which implies the estimator proposed in \eqref{eq:siro_direct} is asymptotically stable almost surely.
\eop

\begin{remark}
According to their definitions, $\mo{dim} (\tilde{\boldsymbol{I_A}}) = \mo{dim}(\tilde{\boldsymbol{A}}) = 2dk_w\times 2d(k_w+1)$, $\mo{dim}(\tilde{\boldsymbol{B}}) = 2dk_w\times dk_w$, $\mo{dim}(\bs{C}) = k_w\times 2d(k_w+1)$, $\mo{dim} (\tilde{\boldsymbol{I}}_A^{\top} \bs{P}\tilde{\boldsymbol{I}}_A) = \mo{dim}(\tilde{\boldsymbol{A}}^{\top}\bs{Q}\tilde{\boldsymbol{A}}) = \mo{dim}(\tilde{\boldsymbol{C}}^{\top}\bs{R}\tilde{\boldsymbol{C}}) = 2d(k_w+1)\times 2d(k_w+1)$. The dimension of ${\boldsymbol{E}}_x$ is $(4d+1)k_w\times 2d(k_w+1)$. Similarly, the dimension of ${\boldsymbol{E}}_{\theta}$ is $(4d+1)k_w\times (3d+1)k_w$. The dimension of $\bs{W}$ is $(4d+1)k_w\times (4d+1)k_w$. Therefore, the dimension of $\bs{W_E}$ is $2d(k_w+1)\times 2d(k_w+1)$. As expected, for a fixed dimension of the state vector, the dimension of $\bs{W_E}$ increases linearly with the sliding window size.
\end{remark}

This least-square estimation problem has a hyperlinear complexity regarding the dimension of $\bs{W_E}$. Therefore, proper dimension reduction of $\bs{W_E}$ is preferred. For this purpose, a polynomial fitting approach is further introduced as follows to enhance computational efficiency.

\begin{lemma}
\label{lem:fiiting}
 For a flying robot with bounded inputs $\bs{u} = [\bs{u}_1, \bs{u}_2, \cdots, \bs{u}_d]$, in a small time interval $[t_i, t_f]$, given an $\epsilon_f$, there exist a $k_t$-th order polynomial $\tilde{\boldsymbol{x}}_i=\sum_{j=0}^{k_t}\alpha_{ij}(t-t_i)^j$ which fits the true trajectories $\bs{x}_i$, such that $\sup (\|\tilde{\boldsymbol{x}}_i-\boldsymbol{x}_i\|)<\epsilon_f$.
\end{lemma}

\textit{Proof}: Regarding \eqref{eq:modelwithdrag} and \eqref{eq:dragmodelwithflow}, the velocity evolution can be formulated as
	\begin{equation}
		\bs{v}_{i}(t) = \boldsymbol{v}_{i0}e^{-\bs{\mu}_{ii} t}+\int_{0}^{t}e^{-\bs{\mu}_{ii} \tau}\bs{u}_i(t-\tau)\mo{d}\tau
		\label{eq:vevolution}
	\end{equation}
where $\bs{v}_{i0}$ is the initial value of $\bs{v}_i$.
	
With Taylor's Expansion, we can rewrite \eqref{eq:vevolution} as
\begin{equation}
    \bs{v}_{i}(t) = f_{\bs{v}_i}(t,t^2,\cdots,t^{k_t}) + R_{\bs{v}_i}
\end{equation}
which contains the following as remainder:
	\begin{equation}
		R_{\bs{v}_i} = \boldsymbol{v}_{i0}h(t)+\int_{0}^{t}h(\tau) u(t-\tau)\mo{d}\tau
	\end{equation}
where $h(t)=\frac{e^{-\bs{\mu}_{ii}\xi t}}{(k_t+1) !}(\bs{\mu}_{ii} t)^{k_t+1}$, and $\xi$ is a number in $(0,1)$.

Denoting $\bar{t}=t_f-t_i$, the maximum value of $h$ should be $M_h=h(\bar{t})$ if $\bar{t} \leq \frac{k_t+1}{\bs{\mu}_{ii}}$, and $M_h=h(\frac{k_t+1}{\bs{\mu}_{ii}})$ if $\bar{t} > \frac{k_t+1}{\bs{\mu}_{ii}}$. Further, denoting the upper bound of the input as $M_u$, i.e., $|\bs{u}_i(t)|<M_u $ for any $ t \in [0,\bar{t}]$, we have
\begin{equation}
 		R_{\bs{v}_i} \leq \boldsymbol{v}_{i0}M_h+M_u\int_0^tM_h \mo{d} \tau =\boldsymbol{v}_{i0}M_h+M_uM_ht
\end{equation}

Integrating the remainder $R_{\bs{v}_i}$, we have the corresponding expansion remainder of the position
\begin{equation}
 R_{\bs{p}_i}
		 \leq M_h\int_0^{\bar{t}}\left(\boldsymbol{v}_{i0}+M_u\tau\right)d\tau
		 = M_h(\boldsymbol{v}_{i0}\bar{t}+\frac{1}{2}M_u\bar{t}^2)
\end{equation}

According to its definition, when $\bar{t}$ is small, $M_h$ monotonically decreases with $k_t$. Therefore, given an $\epsilon_f$, one can always find a significant large $k_t$, such that $\|R_{\bs{p}_i}\|
		 \leq \|M_h(\boldsymbol{v}_{i0}\bar{t}+\frac{1}{2}M_u\bar{t}^2)\|\leq\epsilon_f$.
\eop
\begin{remark}
 Regarding Lemma \ref{lem:fiiting}, when $\bs{\mu}_{ii}t$ is not significant larger than 1, with a fourth order fitting, i.e., $k_t=4$, we can obtain a significant small $M_h\approx 0.01$.
\end{remark}

On this basis, we then approximate the trajectory in a time horizon using $k_t$th order polynomials, that is, $\bs{x}_{j}(t)=\sum_{i=0}^{k_t}(t-t_0)^i\alpha_{ji}\triangleq \sum_{i=0}^{k_t} \bar{t}^i \alpha_{ji}$. The state vector $\bs{x}$ can be then further approximated as ${\bs{x}}=\bs{T}\bs{\alpha}$, where
\begin{equation}
\small
\begin{array}{ll}
\bs{\alpha}=[\bs{\alpha}_{1}^{\top}, \bs{\alpha}_{2}^{\top},\cdots,\bs{\alpha}_{2d}^{\top}]^{\top},&\bs{\alpha}_{j}=[\alpha_{j0}, \alpha_{j1}, \cdots, \alpha_{jk_{t}}]^{\top}\\
\bs{T}=[\bs{T}_{0}^{\top}, \bs{T}_{1}^{\top}, \cdots,  \bs{T}_{k}^{\top}]^{\top},& \bs{T}_{i}  = \diag(\underbrace{\bar{\bs{t}}_{i},\bar{\bs{t}}_{i},\cdots,\bar{\bs{t}}_{i}}_{2d\times \bar{\bs{t}}_{i}}) \vspace{-0.5cm}
\\
 \bar{\bs{t}}_{i} =[1,\bar{t}_{i},\bar{t}^2_{i},\cdots,\bar{t}^{k-1}_i] & \\
\end{array}
\end{equation}

In this way, the optimal estimation problem \eqref{eq:batchls} can be re-formulated as
\begin{equation}
\hat{\bs{\alpha}}=\arg\min_{\bs{\alpha}} J_{\bs{\alpha}} = (\bs{E_x}\bs{T}\bs{\alpha} - \bs{E_{\theta}}\bs{\theta})^{\top}\bs{W}(\bs{E_x}\bs{T}\bs{\alpha} - \bs{E_{\theta}}\bs{\theta})
\label{eq:estpoly}
\end{equation}

Denoting $\bs{E_T}=\boldsymbol{E_xT}$ and taking derivative of $J_{\bs{\alpha}}$, we can obtain
$\frac{\partial J_{\bs{\alpha}}}{\partial \bs{\alpha}} = \bs{E_T}^{\top} \bs{W}(\bs{E_T}\bs{\alpha}-\bs{E_{\theta}}\bs{\theta})$. A best estimation $\hat{\bs{\alpha}}$ is then obtained as
\begin{equation}\label{eq:alphaest}
  \hat{\bs{\alpha}}=(\bs{E_T}^{\top}\bs{W}\bs{E_T})^{-1}\bs{E_T}^{\top}\bs{W}\bs{E_{\theta}}\bs{\theta}
\end{equation}

\begin{corollary}
\label{col:wtinvert}
 Denoting $\bs{W_T} \triangleq \ts{\bs{E_T}}\bs{W}\bs{E_T}$, $\bs{W_T}$ is always invertible. If the original system is observable, the invertibility of $\bs{W_T}$ does not depend on the choice of $\bs{P}$.
\end{corollary}

\proof Based on the definition, the following relationship $\bs{W_T} = \ts{\bs{T}}\bs{W_E}\bs{T}$ establishes. As the matrix $\bs{T}$ has full column rank, $\bs{W_T}$ is still positive definite. Hence, one can obtain the same conclusion as that in Lemma \ref{lm:invertwe} and Corollary \ref{corol:weinvert}.
\eop

\begin{remark}
 As aforementioned, the dimension of $\bs{W_E}$ is $2d(k_w+1)\times 2d(k_w+1)$. For a $k_t$-th order polynomial fitting, the dimension of $\bs{T}$ is $2d(k_w + 1)\times 2d(k_t+1)$. Accordingly, the dimension of $\bs{W_T}$ is reduced to $2d(k_t+1)\times 2d(k_t+1)$. Therefore, the computational complexity is well reduced, especially when the ratio $k_w/k_t$ is big.
\end{remark}

\begin{theorem}
  The estimator proposed in \eqref{eq:estpoly} is asymptotically stable almost surely for the system \eqref{eq:discretemodel}.
\end{theorem}

\proof
Following the proof of Theorem \ref{th:siro_org} and replacing $\bs{W_E}$ with $\bs{W_T} = \ts{\bs{T}}\bs{W_E}\bs{T}$, we can obtain $\rho(\bs{W_T}^{-1}\tbs{T}\tilde{\boldsymbol{I}}_A^{\top} \bs{P}^{-1}\tilde{\boldsymbol{I}}_A\bs{T})< 1$ almost surely, and the estimator's convergence is thereby guaranteed.
\eop

In practice, the sampling frequency of the IMU is commonly higher than that of the UWB. That is, between each two ranging measurements $\ti{\bar{r}}{k}$ and $\ti{\bar{r}}{k+1}$, there are $\ell$ acceleration measurements denoted as $\ti{\bs{a}^{\top}}{k\cdot1}, \ti{\bs{a}^{\top}}{k\cdot2},\cdots, \ti{\bs{a}^{\top}}{k\cdot\ell}\ (\ell \in \mathbb{N}_{>1})$. If we want to use the IMU measurements sufficiently, a pre-integration approach can be introduced as follows. First, we re-formulate the discrete model of \eqref{eq:discretemodel} as
\begin{equation}
 \left\{\begin{array}{l}\ti{\boldsymbol{x}}{k+1}=\bs{A}^{\ell} \ti{\boldsymbol{x}}{k}+ \sum_{j=1}^{\ell}\bs{A}^{\ell-j}\bs{B} \ti{\boldsymbol{u}}{k\cdot j} \\ \ti{y}{k+1}=\bti{C}{k+1}\ti{\boldsymbol{x}}{k+1}\end{array}\right.
 \label{eq:discretemodelsum}
\end{equation}

In this way, $\boldsymbol{\theta}$ is modified as
$\boldsymbol{\theta}=[\ti{\hat{\bs{x}}^{-\top}}{0}, \ti{\hat{\bs{x}}^{-\top}}{1}, \cdots,$ $\ti{\hat{\bs{x}}^{-\top}}{k_w-1}, \ti{\bs{a}^{\top}}{0\cdot1}, \ti{\bs{a}^{\top}}{0\cdot2},\cdots, \ti{\bs{a}^{\top}}{(k_w-1)\cdot\ell}, \ti{\bar{r}}{1}, \ti{\bar{r}}{2},\cdots, \ti{\bar{r}}{k_w}]^{\top}$. This pre-integration procedure also affects the expressions $\tilde{\boldsymbol{A}}$ and $\tilde{\boldsymbol{B}}$, which are explicitly shown in Appendix \ref{app:formab}. With such a modification, the optimal estimation problem can be thereafter re-formulated based on \eqref{eq:alphaest} accordingly.

With the pre-integration technique, one may also increase the time span of the estimation with a fixed $k_w$. That is, assuming the UWB and IMU sampling frequencies are the same, we can directly downsample the range measurement by a factor of $\ell$, and the estimation window can be spanned by $\ell$ times without extra computational time cost.



\subsection{Extended to the SRIFO}
\label{sec:firo}
As formulated in \eqref{eq:dragmodelwithflow}, if an optical flow sensor is integrated for velocity measurement, the output is re-defined as $\bs{y}_f=[\frac{1}{2}\ts{\bs{p}}\bs{p}, \boldsymbol{v}_{f}^{\top}]^{\top}$. The observability of this new system is guaranteed by the following theorem.

\begin{theorem}
\label{th:srifo_ob}
The SRIFO system described with \eqref{eq:dragmodelwithflow} is locally weakly observable almost surely.
\end{theorem}

\proof The observability matrix of the SRIFO system is
\begin{equation}
\label{eq:obmatrix_srifo}
\mathcal{O}=\left[\begin{array}{cc}
\boldsymbol{p}^{\top} & \boldsymbol{0}_{1 \times d} \\
\boldsymbol{0}_{d } & \boldsymbol{I}_{d } \\
\boldsymbol{v}^{\top} & \boldsymbol{p}^{\top} \\
\boldsymbol{0}_{d } & -\boldsymbol{\mu}\boldsymbol{I}_{d} \\
(\boldsymbol{u}-\boldsymbol{\mu} \boldsymbol{v})^{\top} & (2 \boldsymbol{v}-\boldsymbol{\mu} \boldsymbol{p})^{\top} \\
\cdots & \cdots \\
\end{array}\right]
\end{equation}
the detailed derivation of which is presented in \apref{ap:lie_srifo}.

Similar to the proof of Theorem \ref{th:obsiro}, denoting $\bs{\chi_{\mu}} = [{\bs{p}},{\bs{v}},{\bs{u}}-\bs{\mu}\bs{v}]^{\top}$, $ \mathrm{rank}(\mathcal{O})< 2d \iff \bs{\chi_{\mu}}_i \equiv \alpha_j \bs{\chi_{\mu}}_j+\alpha_k \bs{\chi_{\mu}}_k, \bs{\mu}_i \equiv \bs{\mu}_j \equiv \bs{\mu}_k$ ($i, j, k\in[1,d]$). Therefore, the observability matrix \eqref{eq:obmatrix_srifo} is full rank almost surely, which implies that the SRIFO system is locally weakly observable.
\eop

\begin{corollary}
\label{cl:ob_srifo}
For an SRIFO system excluding drag forces, it is still locally weakly observable almost surely.
\end{corollary}

\proof Setting $\boldsymbol{\mu}=\bs{0}$ in \eqref{eq:obmatrix_srifo}, and denoting $\bs{\chi_{0}} = [{\bs{p}},{\bs{v}},{\bs{u}}]^{\top}$, one may obtain the same conclusion as that in the proof of Theorem \ref{th:srifo_ob}.
\eop

Corollary \ref{cl:ob_srifo} implies the aerial drag effects play a much less important role in the SRIFO compared to the SRIO, because the velocity measurement is available via the optical flow sensor. Although there are differences, the SRIO can be seamlessly extended to the SRIFO. Specifically, when the velocity measurement is included, the process model characterized with matrices $\bs{A}$ and $\bs{B}$ remains the same as that in \eqref{eq:discretemodel}, whereas the observation model is re-written as $\tli{\bs{y}}{f}{k}=\tli{\boldsymbol{C}}{f}{k}\bti{x}{k}$, with $\tli{\bs{C}}{f}{k}=\diag ( \bti{p}{k}/\|\bti{p}{k}\| , \bs{I}_{d})$.
We then replace $\ti{\boldsymbol{C}}{k}$ and $\ti{\bar{r}}{i}$ with $\tli{\boldsymbol{C}}{f}{k}$ and $\tli{\bar{\bs{r}}}{f}{i}=[\ti{\bar{r}}{i}, \tli{\bar{\bs{v}}}{f}{i}^{\top}]^{\top}$ in \eqref{eq:alphaest}, respectively. In this way, the optimal estimation of the SRIFO is formulated as
\begin{equation}
\hat{\bs{\alpha}}_f=(\bs{E_{Tf}}^{\top}\bs{W}_f\bs{E_{Tf}})^{-1}\bs{E_{Tf}}^{\top}\bs{W}_f\bs{E_{\theta f}}\bs{\theta}_f
\end{equation}
 where ${\boldsymbol{\theta}}_f$ is re-defined as ${\boldsymbol{\theta}}_f=[\ti{\hat{\bs{x}}^{-\top}}{0}, \ti{\hat{\bs{x}}^{-\top}}{1}, \cdots,\ti{\hat{\bs{x}}^{-\top}}{k_w-1}$, $\ti{\bs{a}^{\top}}{0}, \ti{\bs{a}^{\top}}{1},\cdots, \ti{\bs{a}^{\top}}{k-1}, \tli{\bs{\bar{r}}^{\top}}{f}{1}, \tli{\bs{\bar{r}}^{\top}}{f}{2},\cdots, \tli{\bs{\bar{r}}^{\top}}{f}{k}]^{\top}$, and all related matrices $\bs{M}_{\cdot f}$ are modified according to the observation model to guarantee conformability.

\begin{remark}
 In this way, the dimension of ${\boldsymbol{C}}_f$ increases to $4k_w$. The dimensions of ${\boldsymbol{E}}_{xf}$ and ${\boldsymbol{E}}_{\theta f}$ increase to $(4d+4)k_w\times 2d(k_w+1)$ and $(4d+4)k_w\times (3d+4)k_w$, respectively. However, as the state vector does not change, the dimension of $\bs{W_{Tf}}\triangleq\bs{E_{Tf}}^{\top}\bs{W}_f\bs{E_{Tf}}$ is the same as that of $\bs{W_{T}}$.
\end{remark}

\subsection{Fault Tolerance}

The UWB may fail when obstacles exist between the ranging tag and anchor, and the optical flow may fail when the illumination and texture conditions are not satisfactory. In this section, we further demonstrate that our proposed estimator is tolerant to these failures.

As we have discussed in Lemma \ref{lm:invertwe}, one can always obtain a unique state estimation independent from the observability conditions of the original system. Equivalently, the numeric stability of the SRIFO does not depend on the form or rank of the output matrix $\bs{C}_f$. In such a case, when some sensor fails, we can directly obtain a resilient estimation by setting the corresponding measurement covariance to infinity. Mathematically, we denote the measurement covariance in the SRIFO as $\bs{R}_f =\mo{diag} (R_r, \bs{R}_{of})$, where $R_r$ is the covariance of the UWB and $\bs{R}_{of}$ is the measurement covariance of the optical flow, supposing measurements from the UWB and optical flow are independent of each other. On this basis, when the UWB fails, one can set $R_r=\infty$ ($R_r^{-1}=0$), and when the optical flow fails, one can set $\bs{R}_{of}=\bs{\infty}$ ($\bs{R}_{of}^{-1}=\bs{0}$). In particular, when we set $\bs{R}_{of}^{-1}=\bs{0}$, the velocity measurement is excluded, and the SRIFO is degraded to SRIO algorithm. Therefore, one can implement the SRIFO as a general framework, and the ranging and/or optical flow can be excluded in real time according to the implementation or performance of the corresponding sensor. Finally, if the covariance of $R_r$ and $\bs{R}_{of}$ can be obtained in real-time, one can also dynamically allocate them to improve the performance of the SRIFO algorithm.

\begin{figure}
  \vspace{0.1 cm}
  \centering
  \includegraphics[width=0.98\columnwidth]{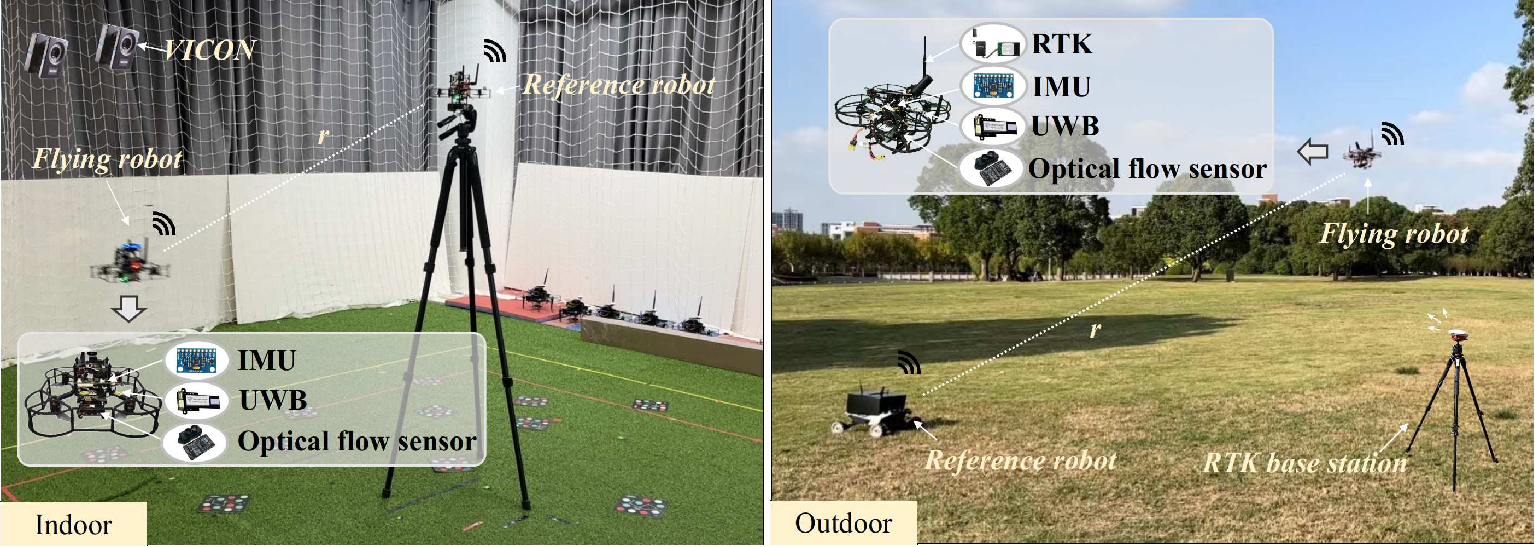}
  \caption{Experimental setup in indoor and outdoor environments}\label{fig:testbed}
\end{figure}

\begin{figure}
  \centering
  \subfigure[Acceleration]{\includegraphics[width=0.48\columnwidth]{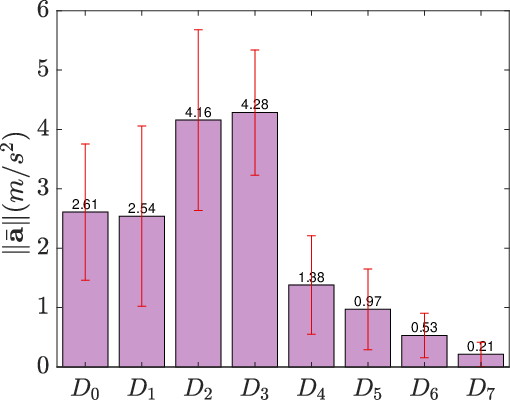}}
  \subfigure[Velocity]{\includegraphics[width=0.49\columnwidth]{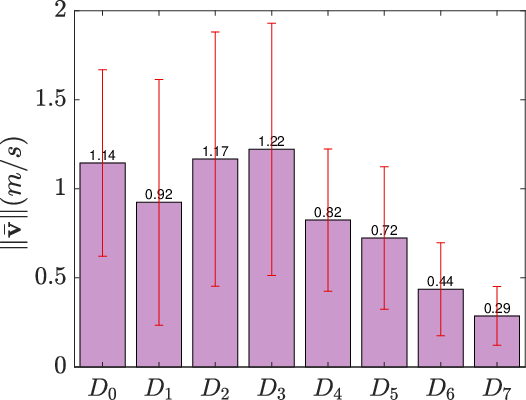}}
  \caption{The acceleration and velocity distributions in indoor experiments}\label{fig:avgvelacc}
\end{figure}

\begin{figure}
  \centering
  \includegraphics[width=\columnwidth]{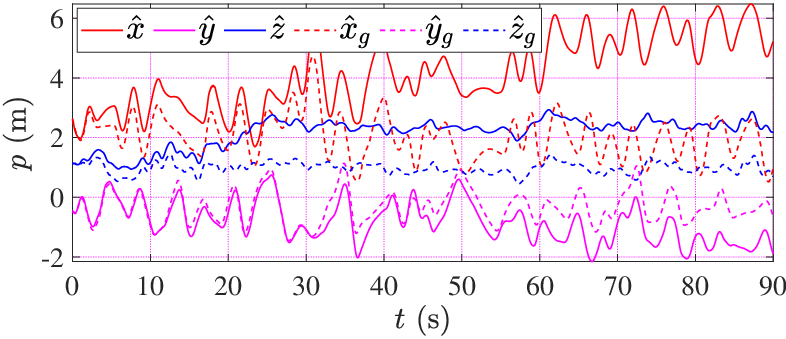}
  \caption{Position estimation with the velocity measured from the optical flow}\label{fig:flow_estimate}
\end{figure}

\section{Experiments}

To efficiently assess the effectiveness of the proposed approach, experiments are conducted in both indoor and outdoor scenarios.

\subsection{Experimental Setup}
In all experiments, two custom developed quadrotor flying robots controlled by the open-source Pixhawk\textsuperscript{\textregistered} firmware are steered to fly in random trajectories. An UP Core\textsuperscript{\textregistered} plus computing board mounted Intel Atom x7 (four cores, 1.8 GHz) is adopted as the onboard computer. The IMU (model MPU9250) embedded on the Pixhawk\textsuperscript{\textregistered} is adopted for acceleration measurement. The Nooploop\textsuperscript{\textregistered} UWB radio (model LinkTrack LTPS\textsuperscript{\textregistered}) is selected for range measurement. The optical flow sensor (model anotc v4.0) is implemented for the verification of the SRIFO.

\begin{figure*}[!t]
  \centering
  \subfigure[$k_t=2$]{\includegraphics[width=0.49\columnwidth]{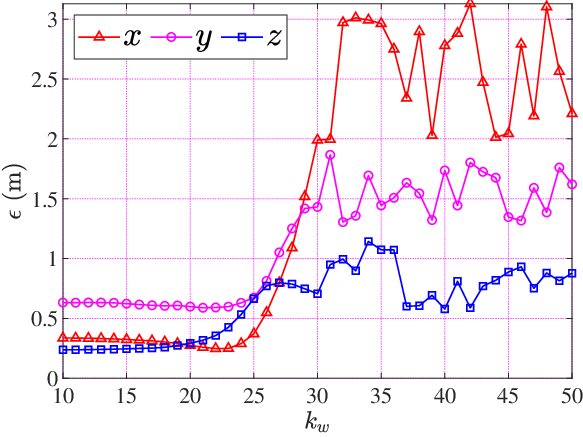}}
  \subfigure[$k_t=3$]{\includegraphics[width=0.49\columnwidth]{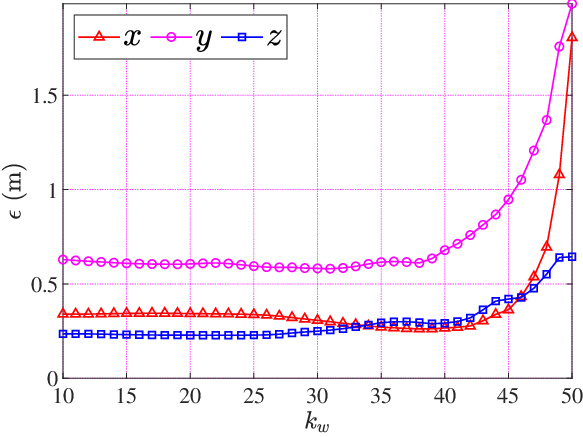}}
  \subfigure[$k_t=4$]{\includegraphics[width=0.49\columnwidth]{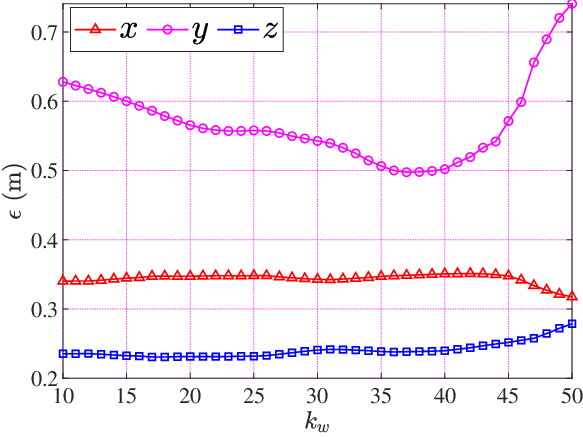}}
  \subfigure[$k_t=6$]{\includegraphics[width=0.49\columnwidth]{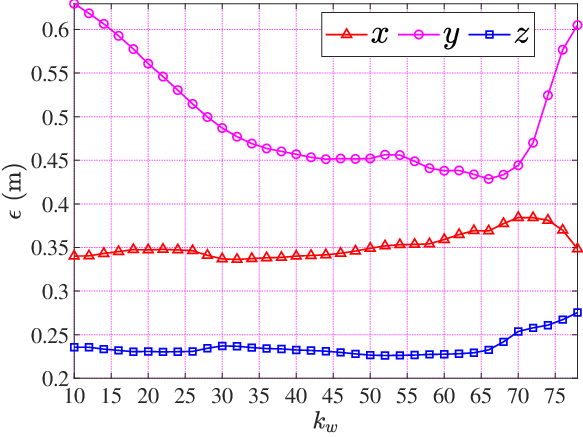}}
  \caption{The MAEs of the SRIO with different fitting order $k_t$ and window size $k_w$}\label{fig:varyingktkw}
\end{figure*}

In the indoor environment, a VICON\textsuperscript{\textregistered} motion capture system is used to collect ground truth state. The motion capture system runs at a frequency of 200 Hz. It estimates the state of the airframe by kinematic fitting the reflective markers which are attached to the aerial platforms. A homogeneous `reference robot' is depolyed to either simulate a fixed anchor or a cooperative robot. In the outdoor environment, the ground truth state is recorded using a real-time kinematic positioning module (model C-RTK 9Ps). It is compatible with Pixhawk\textsuperscript{\textregistered} and provides centimeter-level positioning precision. As robots in outdoor environment have additional payloads of the RTK, they have bigger airframes compared with that in the indoor environment. Apart from this, a heterogeneous ground `reference robot' is deployed to simulate a fixed anchor or a cooperative robot. The configurations of the aforementioned two testbeds are illustrated in \figref{fig:testbed}.

\subsection{Collection of Data}

\begin{figure}
  \centering
   \includegraphics[width=0.88\columnwidth]{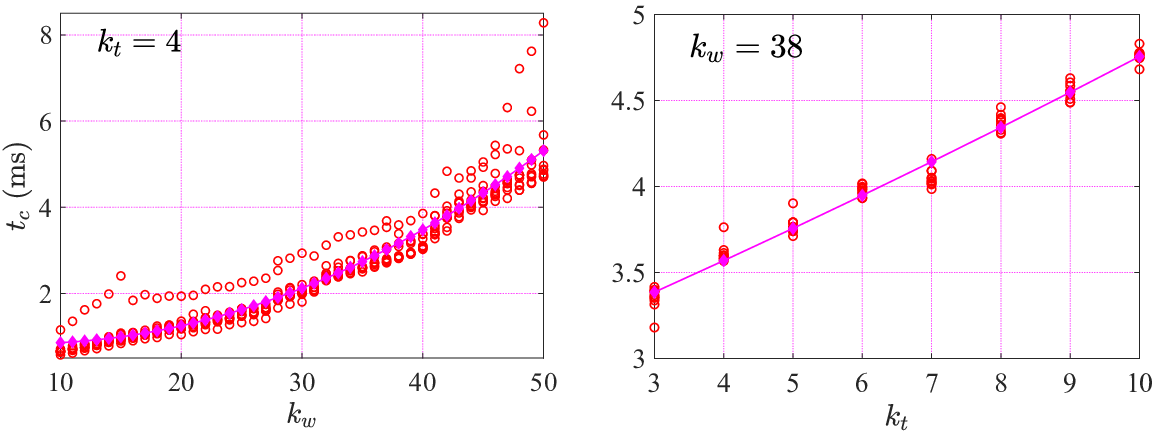}
  \caption{Time cost with different fitting order $k_t$ and window size $k_w$}\label{fig:costime_srio}
\end{figure} 

Because the experimental evaluation mainly focuses on the effectiveness of the state estimation, autonomous flight control is not primarily considered in this work. Therefore, flying robots are manually steered by an operator to fly in random trajectories. We repeated the experiments with different velocities, and the IMU, UWB, and optical flow measurements are collected at a frequency of 25 Hz. We then explore the effectiveness of our estimation approach using the recorded measurements and the ground truth.

Eight datasets, namely $D_0\sim D_7$, are recorded in the indoor testbed. Their acceleration and velocity distributions are illustrated in \figref{fig:avgvelacc}. The characteristics of the raw sensor data are demonstrated in Appendix \ref{app:measurecurve} using a part of measurements in $D_0$ and $D_2$. As expected, noises and/or drifts are observed in all measurements. If only one source of data, e.g., velocity measurement from the optical flow, is utilized for positioning, significant divergences occur as depicted in \figref{fig:flow_estimate}.

Meanwhile, four datasets, $E_0\sim E_3$, are also recorded in the outdoor experiments simulating a normal flight with moderate velocities and accelerations, which are similar to that of $D_4$ and $D_5$. Compared with indoor experiments, the flying robot is steered to fly in a much longer distance in the outdoor environment. In the following discussion, we first explore the effectiveness of the SRIBO using the $D_0\sim D_7$, and then extend the verification to $E_0\sim E_3$.

\subsection{Performance of the SRIO}

 \begin{table}[!b] 
 \centering
 \caption{Estimation errors of SRIO with different fitting order $k_t$}
 \label{tab:siro_err}
   \begin{threeparttable}
 \begin{tabular}{p{1mm}p{2mm}p{3.5mm}p{3.5mm}p{6mm}p{3.5mm}p{3.5mm}p{6mm}p{3.5mm}p{3.5mm}p{3.5mm}}
  \toprule
  \multicolumn{2}{c}{\textbf{Coefs.}} & \multicolumn{3}{c}{\textbf{MAE (m)}} & \multicolumn{3}{c}{\textbf{RMSE (m)}} &\multicolumn{3}{c}{\textbf{SAE (m)}}\\
  \cline{1-2}\cline{3-5} \cline{6-8} \cline{9-11}
  $k_t$ & $k_w$ & ${x}$ & ${y}$ & ${z}$ & ${x}$ & ${y}$ & ${z}$& ${x}$ & ${y}$ & ${z}$ \\
  \midrule
  {4} & 38 & 0.35 & 0.50 & 0.24 & 0.44 & 0.62 & 0.29 & 0.27 & 0.38 & 0.17 \\
  {5} & 46 & 0.34 & 0.45 & 0.22 & 0.43 & 0.57 & 0.27 & 0.26 & 0.35 & 0.16 \\
  {6} & 44 & 0.34 & 0.45 & 0.23 & 0.43 & 0.56 & 0.28 & 0.26 & 0.34 & 0.16 \\
  {7} & 52 & 0.34 & 0.42 & 0.22 & 0.43 & 0.54 & 0.27 & 0.26 & 0.33 & 0.15 \\
  {8} & 52 & 0.34 & 0.41 & 0.22 & 0.42 & 0.53 & 0.27 & 0.26 & 0.33 & 0.16 \\
  \bottomrule
 \end{tabular}
   \end{threeparttable}
\end{table}

 \begin{table}[!b]
 \centering
 \caption{Estimation errors of SRIO using different datasets }
 \label{tab:siro_err_ds3}
   \begin{threeparttable}
 \begin{tabular}{p{6mm}p{3.5mm}p{3.5mm}p{6mm}p{3.5mm}p{3.5mm}p{6mm}p{3.5mm}p{3.5mm}p{3.5mm}}
  \toprule
  {\textbf{Data}} & \multicolumn{3}{c}{\textbf{MAE (m)}} & \multicolumn{3}{c}{\textbf{RMSE (m)}} &\multicolumn{3}{c}{\textbf{SAE (m)}}\\
  \cline{2-4} \cline{5-7} \cline{8-10}
  \textbf{Set} & ${x}$ & ${y}$ & ${z}$ & ${x}$ & ${y}$ & ${z}$& ${x}$ & ${y}$ & ${z}$ \\
  \midrule
  {$D_0$}   & 0.35 & 0.50 & 0.24 & 0.44 & 0.62 & 0.29 & 0.27 & 0.38 & 0.16 \\
  {$D_1$}   & 0.20 & 0.61 & 0.20 & 0.25 & 0.76 & 0.25 & 0.15 & 0.45 & 0.15 \\
  {$D_2$}   & 0.36 & 0.33 & 0.46 & 0.41 & 0.43 & 0.57 & 0.19 & 0.27 & 0.33 \\
  {$D_3$}   & 0.40 & 0.41 & 0.47 & 0.48 & 0.57 & 0.56 & 0.26 & 0.39 & 0.30 \\
  {$D_4$}   & 0.34 & 0.46 & 0.31 & 0.45 & 0.60 & 0.39 & 0.30 & 0.39 & 0.24 \\
  {$D_5$}   & 0.36 & 0.90 & 0.43 & 0.47 & 1.03 & 0.49 & 0.29 & 0.50 & 0.23 \\
  {$D_6$}   & 0.61 & 0.57 & 0.29 & 0.97 & 0.76 & 0.40 & 0.75 & 0.50 & 0.28 \\
  {$D_7$}   & 0.64 & 0.46 & 0.60 & 1.00 & 0.74 & 0.64 & 0.76 & 0.58 & 0.22 \\
  \bottomrule
 \end{tabular}
   \end{threeparttable}
\end{table}

\begin{figure}
  \centering
  \includegraphics[width=0.99\columnwidth]{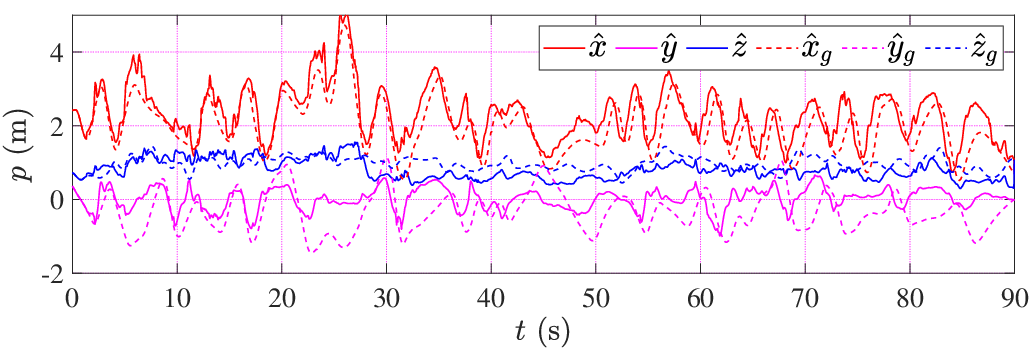}
  \caption{Typical position estimation result using the SIRO (dataset $D_0$)}\label{fig:bag5_siro_fit_4th}
  \label{bag5_siro_fit_4th}
\end{figure} 

We first tune the parameter of the SRIO in a trial and error approach, and all parameter values are listed in Appendix \ref{app:sriopar}. Using measurements in first 40 s from the dataset $D_0$, we investigate the performance of the SRIO with varying fitting order $k_t$ and windows size $k_w$. As shown in \figref{fig:varyingktkw}, the performance of the SRIO steadily improves with increasing window size $k_w$. However, when the fitting order is small, especially when $k_t\leq 3$, the performance rapidly deteriorates when $k_w$ is bigger than a certain size. Apparently, a small fitting order is not able to effectively fit a long interval.

More detailed performance indices of the SRIO with optimal combination of $k_t$ and $k_w$ regarding \figref{fig:varyingktkw} are also illustrated in \tabref{tab:siro_err}. The mean absolute error (MAE), root mean squre error (RMSE), and standard deviation of absolute error (SAE) almost remain the same when $k_t\geq 4$. Subsequently, we further investigate the time efficiency of SRIO with different combination of $k_t$ and $k_w$, and the results are illustrated in \figref{fig:costime_srio}. It can be seen that the time cost is approximately linearly correlated with the fitting order $k_t$, and approximately quadratically correlated with the window size $k_w$.  In view of those results, considering the performance and time efficiency, this work selects $k_t=4$ and $k_w=38$. With such a selection, the estimation results of the SRIO using dataset $D_0$ are depicted in \figref{bag5_siro_fit_4th}.

\begin{table}[!b]
\centering
 \caption{Performance of SRIO without considering aerial drag effects}
 \label{tab:sriowodrags}
   \begin{threeparttable}
\begin{tabular}{p{6mm}p{4mm}p{4mm}p{5mm}p{4mm}p{4mm}p{5mm}p{4mm}p{4mm}p{4mm}}
  \toprule
  {\textbf{Data}} & \multicolumn{3}{c}{\textbf{MAE (m)}} & \multicolumn{3}{c}{\textbf{RMSE (m)}} &\multicolumn{3}{c}{\textbf{SAE (m)}}\\
  \cline{2-4} \cline{5-7} \cline{8-10}
  {\textbf{Set}} & ${x}$ & ${y}$ & ${z}$ & ${x}$ & ${y}$ & ${z}$& ${x}$ & ${y}$ & ${z}$ \\
  \midrule
  {$D_0$}   & 1.18 & 0.79 & 1.20 & 1.44 & 0.95 & 1.39 & 0.83 & 0.53 & 0.69 \\
  {$D_1$}   & 0.99 & 0.60 & 1.11 & 1.20 & 0.77 & 1.22 & 0.68 & 0.49 & 0.49 \\
  {$D_2$}   & 0.41 & 0.62 & 0.56 & 0.50 & 0.83 & 0.68 & 0.29 & 0.56 & 0.38 \\
  {$D_3$}   & 0.95 & 0.68 & 0.51 & 1.32 & 1.00 & 0.67 & 0.91 & 0.73 & 0.43 \\
  {$D_4$}   & 0.44 & 1.06 & 0.46 & 0.54 & 1.16 & 0.54 & 0.32 & 0.48 & 0.29 \\
  {$D_5$}   & 0.37 & 2.08 & 0.64 & 0.46 & 2.26 & 0.81 & 0.27 & 0.89 & 0.50 \\
  {$D_6$}   & 0.79 & 0.69 & 0.15 & 1.18 & 0.81 & 0.22 & 0.88 & 0.42 & 0.16 \\
  {$D_7$}   & 0.31 & 1.07 & 0.57 & 0.39 & 1.50 & 0.72 & 0.22 & 1.06 & 0.44 \\
  \bottomrule
 \end{tabular}
   \end{threeparttable}
\end{table}

\begin{figure}
  \centering
  \includegraphics[width=0.98\columnwidth]{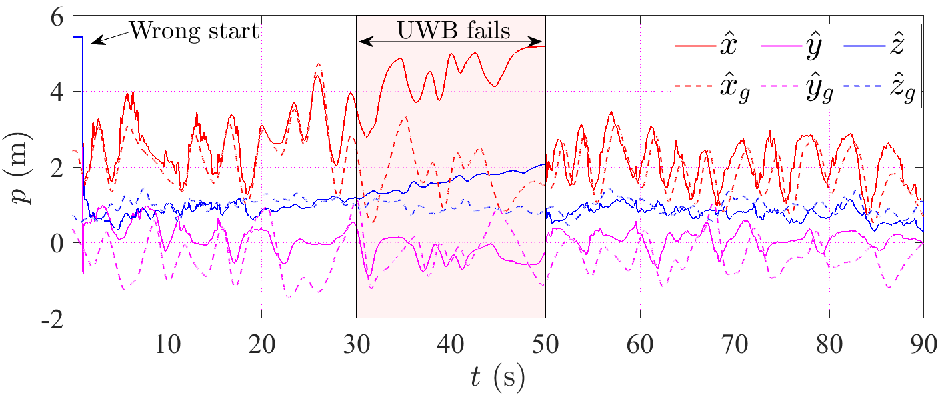}
  \caption{Fault tolerant performance of the SIRO (dataset $D_0$)}\label{fig:wronginit}
\end{figure} 

\begin{figure}
  \centering
  \includegraphics[width=0.98\columnwidth]{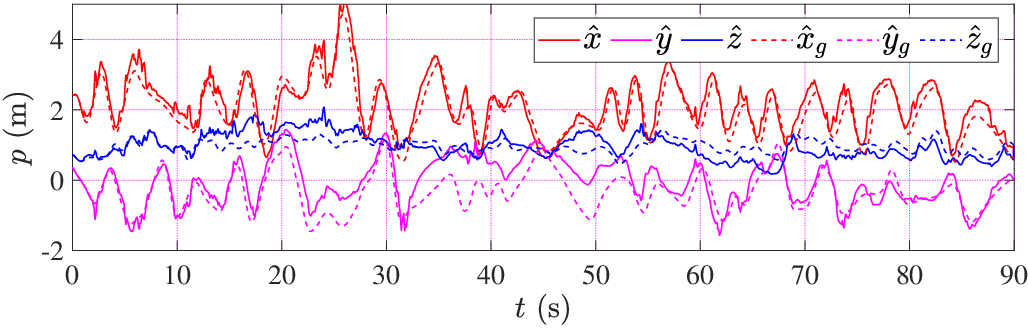}
  \caption{Typical position estimation result using the SRIFO (dataset $D_0$)}\label{fig:bag5_srifo}
\end{figure} 

We then evaluate the performance of SRIO using different datasets. The results are provided in Table \ref{tab:siro_err_ds3}. The estimation precision using $D_1\sim D_7$ is all similar to that of $D_0$, although slight differences are observed owning to different acceleration and velocity distributions regarding \figref{fig:avgvelacc}. Specifically, the accelerations as well as velocities in datasets $D_4\sim D_7$ are smaller than that of $D_0\sim D_3$, the corresponding estimation errors increases evidently. Nonetheless, the estimation performance remains stable even in handling the quasi-static case $D_7$. This means the performance of the proposed SRIO is not significantly affected by the excitation pattern of flying robots. When compared to the results reported by existing work \cite{dongTrajectoryEstimationFlying2022,cossetteRelativePositionEstimation2021}, the SRIO proposed in this work also demonstrates a steady performance with higher precision and efficiency.


Thereafter, we investigate the significance of our refined model considering aerial drag effects. As shown in Table \ref{tab:sriowodrags}, when the drag effects are neglected, even if we still use the DWE proposed by this work, the estimation performance deteriorates in most cases. This confirms that our refined model enhances the observability of this SRIO system.

Finally, utilizing dataset $D_0$, we test the resilience of the SRIO by simulating a wrong initialization and a UWB failure in a certain horizon. As demonstrated in \figref{fig:wronginit}, even if we exaggeratedly wrongly guess an initial state as $\hat{\bs{x}}_0= 2\bar{r}_0\times\bs{1}_{6\times 1}$, the estimation can still quickly converge. Apart from this, we simulate a UWB failure in the time horizon $[30,50]$. Although the estimation divergences when this failure occurs, the estimation quickly converges when the UWB is available after $t=50$ s.

 \begin{table}[b]
 \centering
 \caption{Estimation error of SRIFO}
 \label{tab:firo_err_ds3}
   \begin{threeparttable}
 \begin{tabular}{p{6mm}p{3.5mm}p{3.5mm}p{6mm}p{3.5mm}p{3.5mm}p{6mm}p{3.5mm}p{3.5mm}p{3.5mm}}
  \toprule
  {\textbf{Data}} & \multicolumn{3}{c}{\textbf{MAE (m)}} & \multicolumn{3}{c}{\textbf{RMSE (m)}} &\multicolumn{3}{c}{\textbf{SAE (m)}}\\
  \cline{2-4} \cline{5-7} \cline{8-10}
  {\textbf{Set}} & ${x}$ & ${y}$ & ${z}$ & ${x}$ & ${y}$ & ${z}$& ${x}$ & ${y}$ & ${z}$ \\
  \midrule
  {$D_0$}   & 0.27 & 0.30 & 0.23 & 0.33 & 0.39 & 0.31 & 0.19 & 0.25 & 0.21 \\
  {$D_1$}   & 0.24 & 0.34 & 0.22 & 0.30 & 0.43 & 0.29 & 0.18 & 0.27 & 0.19 \\
  {$D_2$}   & 0.31 & 0.85 & 0.86 & 0.41 & 1.07 & 1.16 & 0.26 & 0.64 & 0.78 \\
  {$D_3$}   & 0.36 & 0.75 & 0.42 & 0.49 & 0.99 & 0.54 & 0.32 & 0.65 & 0.35 \\
  {$D_4$}   & 0.24 & 0.34 & 0.07 & 0.28 & 0.37 & 0.10 & 0.15 & 0.14 & 0.07 \\
  {$D_5$}   & 0.24 & 0.39 & 0.06 & 0.34 & 0.52 & 0.07 & 0.24 & 0.34 & 0.04 \\
  {$D_6$}   & 0.25 & 0.25 & 0.24 & 0.31 & 0.32 & 0.32 & 0.18 & 0.20 & 0.22 \\
  {$D_7$}   & 0.18 & 0.20 & 0.18 & 0.24 & 0.34 & 0.23 & 0.15 & 0.27 & 0.14 \\
  \bottomrule
 \end{tabular}
   \end{threeparttable}
\end{table}

\subsection{Performance of the SRIFO}

\begin{figure}
  \centering
  \includegraphics[width=0.98\columnwidth]{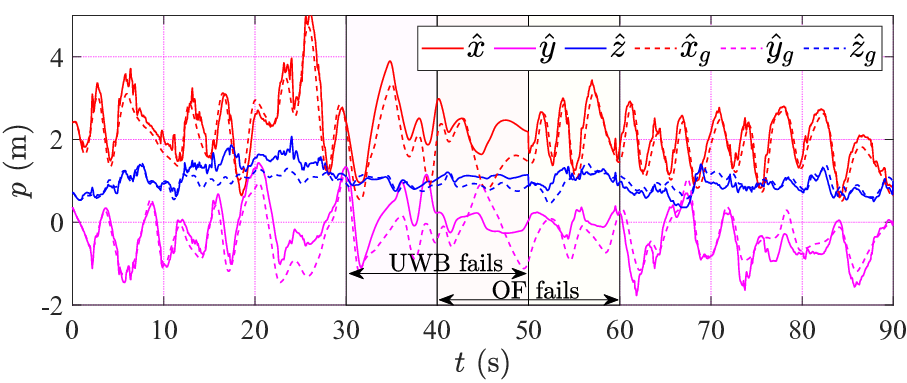}
  \caption{Fault tolerant performance of the SRIFO (dataset $D_0$)}\label{fig:bag5_sfiro_ft}
\end{figure} 

To explore the generality of the SRIBO framework, we further introduce velocity measurements from the optical flow to perform the SRIFO. The parameters of the SRIFO are selected the same as that of the SRIO except $\bs{R}^{-1}=\bs{I}_4$. A typical estimation result using dataset $D_0$ is illustrated in \figref{fig:bag5_srifo}. As additional velocity measurements is included, the estimation performance is much better than that of the SRIO. Different datasets are also adopted for a further verification. As demonstrated in Table \ref{tab:firo_err_ds3}, the estimation performance is significantly enhanced in most trials compared to that shown in Table \ref{tab:siro_err_ds3}. However, there are exceptional cases $D_2$ and $D_3$, the performance of which is even worse than that of the SRIO. This is mainly due to the undesired performance of the optical flow sensor, which is also illustrated in Appendix \ref{app:measurecurve}. This indicates although the optical flow sensor may enhance the estimation performance, it still could be easily affected by the reliability of the velocity measurement, which could affected by the environmental and illuminating conditions. However, as our SRIBO as well as the DWE framework is formulated for general implementations, one can also practically adopt more accurate and robust sensors to enhance the overall performance based on the minimal configuration of SRIO.

Similar to the resilience verification of the SRIO, a UWB failure is simulated at the interval [30,50] and an optical flow failure is simulated at the interval [40,60]. The estimation results are illustrated in \figref{fig:bag5_sfiro_ft}, where `optical flow' is abbreviated as `OF'. Although the estimation diverges when one or two sensors fail to work, it quickly converges when those sensors work again. These characteristics confirms the numerical stability and convergence we have declared in Section \ref{sec:dwe}.

\begin{table}
\centering
 \caption{Performance of the SRIFO excluding aerial drag effects}
 \label{tab:srifowodrags}
   \begin{threeparttable}
 \begin{tabular}{p{6mm}p{4mm}p{4mm}p{5mm}p{4mm}p{4mm}p{5mm}p{4mm}p{4mm}p{4mm}}
  \toprule
  {\textbf{Data}} & \multicolumn{3}{c}{\textbf{MAE (m)}} & \multicolumn{3}{c}{\textbf{RMSE (m)}} &\multicolumn{3}{c}{\textbf{SAE (m)}}\\
  \cline{2-4} \cline{5-7} \cline{8-10}
  {\textbf{Set}} & ${x}$ & ${y}$ & ${z}$ & ${x}$ & ${y}$ & ${z}$& ${x}$ & ${y}$ & ${z}$ \\
  \midrule
  {$D_0$}  & 0.28 & 0.31 & 0.23 & 0.35 & 0.40 & 0.31 & 0.20 & 0.26 & 0.20 \\
  {$D_1$}  & 0.24 & 0.34 & 0.24 & 0.30 & 0.43 & 0.31 & 0.18 & 0.26 & 0.20 \\
  {$D_2$}  & 0.31 & 0.85 & 0.87 & 0.41 & 1.07 & 1.17 & 0.27 & 0.64 & 0.78 \\
  {$D_3$}  & 0.37 & 0.74 & 0.42 & 0.49 & 0.99 & 0.55 & 0.33 & 0.65 & 0.35 \\
  \bottomrule
 \end{tabular}
   \end{threeparttable}
\end{table}

At last, we verify the role of drag effects in the estimation with SRIFO by comparatively testing with datasets $D_0\sim D_3$. Compared with the SRIO, the SRIFO is less affected as shown in Table \ref{tab:srifowodrags}, which confirms our analysis in Section \ref{sec:firo}. Intuitively, the drag effects avoid the divergence of velocity. When velocity measurements from the optical flow are included, the velocity estimation can be effectively corrected. Therefore, the drag effects is not so significant in the modeling and estimation of the SRIFO.

 \begin{table}[t]
 \centering
 \caption{Estimation errors of SRIO using different datasets }
 \label{tab:srio_outdoor}
   \begin{threeparttable}
 \begin{tabular}{p{6mm}p{3.5mm}p{3.5mm}p{6mm}p{3.5mm}p{3.5mm}p{6mm}p{3.5mm}p{3.5mm}p{3.5mm}}
  \toprule
  {\textbf{Data}} & \multicolumn{3}{c}{\textbf{MAE (m)}} & \multicolumn{3}{c}{\textbf{RMSE (m)}} &\multicolumn{3}{c}{\textbf{SAE (m)}}\\
  \cline{2-4} \cline{5-7} \cline{8-10}
  \textbf{Set} & ${x}$ & ${y}$ & ${z}$ & ${x}$ & ${y}$ & ${z}$& ${x}$ & ${y}$ & ${z}$ \\
  \midrule
  {$E_0$}   & 0.82 & 0.53 & 0.37 & 0.97 & 0.66 & 0.46 & 0.51 & 0.40 & 0.27 \\
  {$E_1$}   & 0.28 & 0.84 & 0.48 & 0.34 & 1.06 & 0.57 & 0.18 & 0.64 & 0.32 \\
  {$E_2$}   & 0.26 & 0.86 & 0.41 & 0.30 & 1.06 & 0.50 & 0.16 & 0.62 & 0.30 \\
  {$E_3$}   & 0.45 & 0.80 & 0.47 & 0.55 & 0.98 & 0.58 & 0.32 & 0.56 & 0.34 \\
  \bottomrule
 \end{tabular}
   \end{threeparttable}
\end{table}

\begin{figure}
  \centering
  \includegraphics[width=0.98\columnwidth]{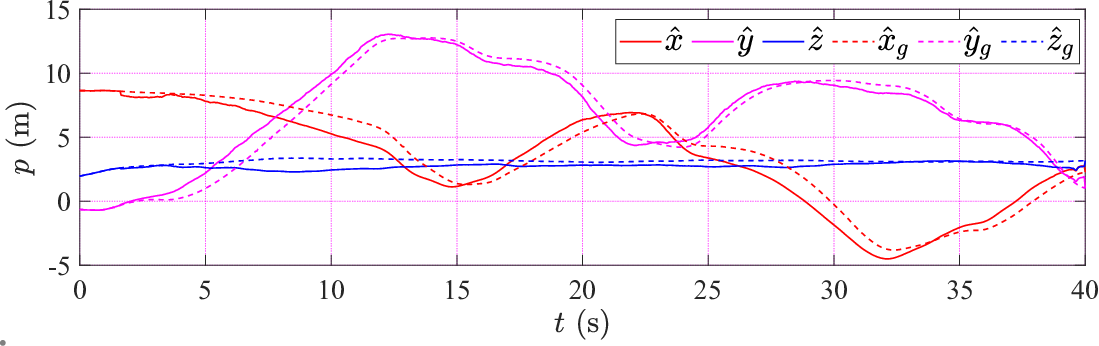}
  \caption{Typical estimation performance of the SRIO (dataset $E_0$)}\label{fig:outdoorsrio}
\end{figure} 
\subsection{Outdoor experiments}

Because we use a different platform in the outdoor environment, the parameters of the SRIBO are re-tuned and listed in Appendix \ref{app:sriopar}.

 \begin{table}[t]
 \centering
 \caption{Estimation errors of SRIO excluding drag effecs }
 \label{tab:siroexdragoutdoor}
   \begin{threeparttable}
 \begin{tabular}{p{6mm}p{3.5mm}p{3.5mm}p{6mm}p{3.5mm}p{3.5mm}p{6mm}p{3.5mm}p{3.5mm}p{3.5mm}}
  \toprule
  {\textbf{Data}} & \multicolumn{3}{c}{\textbf{MAE (m)}} & \multicolumn{3}{c}{\textbf{RMSE (m)}} &\multicolumn{3}{c}{\textbf{SAE (m)}}\\
  \cline{2-4} \cline{5-7} \cline{8-10}
  \textbf{Set} & ${x}$ & ${y}$ & ${z}$ & ${x}$ & ${y}$ & ${z}$& ${x}$ & ${y}$ & ${z}$ \\
  \midrule
  {$E_0$}   & 3.46 & 1.83 & 1.98 & 4.24 & 2.10 & 2.41 & 2.45 & 1.03 & 1.36 \\
  {$E_1$}   & 1.64 & 1.56 & 1.81 & 1.91 & 1.96 & 2.16 & 0.97 & 1.19 & 1.18 \\
  {$E_2$}   & 2.51 & 1.03 & 3.54 & 2.88 & 1.21 & 3.78 & 1.41 & 0.63 & 1.33 \\
  {$E_3$}   & 1.31 & 1.43 & 2.65 & 1.52 & 1.64 & 2.96 & 0.76 & 0.80 & 1.31 \\
  \bottomrule
 \end{tabular}
   \end{threeparttable}
\end{table}

 \begin{table}[t]
 \centering
 \caption{Estimation errors of SRIFO in outdoor environment }
 \label{tab:srifo_outdoor}
   \begin{threeparttable}
 \begin{tabular}{p{6mm}p{3.5mm}p{3.5mm}p{6mm}p{3.5mm}p{3.5mm}p{6mm}p{3.5mm}p{3.5mm}p{3.5mm}}
  \toprule
  {\textbf{Data}} & \multicolumn{3}{c}{\textbf{MAE (m)}} & \multicolumn{3}{c}{\textbf{RMSE (m)}} &\multicolumn{3}{c}{\textbf{SAE (m)}}\\
  \cline{2-4} \cline{5-7} \cline{8-10}
  \textbf{Set} & ${x}$ & ${y}$ & ${z}$ & ${x}$ & ${y}$ & ${z}$& ${x}$ & ${y}$ & ${z}$ \\
  \midrule
  {$E_0$}   & 0.28 & 0.27 & 0.11 & 0.36 & 0.33 & 0.13 & 0.22 & 0.18 & 0.07 \\
  {$E_1$}   & 0.28 & 0.68 & 0.28 & 0.36 & 0.78 & 0.34 & 0.23 & 0.39 & 0.20 \\
  {$E_2$}   & 0.22 & 0.17 & 0.06 & 0.24 & 0.20 & 0.08 & 0.10 & 0.10 & 0.05 \\
  {$E_3$}   & 0.48 & 0.24 & 0.11 & 0.53 & 0.29 & 0.14 & 0.21 & 0.16 & 0.09 \\
  \bottomrule
 \end{tabular}
   \end{threeparttable}
\end{table}

Typical estimation results of the SRIO using dataset $E_0$ are illustrated in \figref{fig:outdoorsrio}. Although handling a case with a much longer distance, the estimation can still closely track the ground truth. Quantitatively, we also estimate the position of the robot regarding different datasets, and results are indicated in Table \ref{tab:srio_outdoor}. Compared with the indoor experiments, the positioning performance deteriorates significantly. This is mainly because unpredicted wind gusts exist in the outdoor environment. As one can expect, such wind gusts will add extra complexities to the dynamics of the flying robot. Nonetheless, the refined model considering aerial drag effects is still significant. As demonstrated in Table \ref{tab:siroexdragoutdoor}, when the aerial drag effects are excluded in the model, the positioning performance deteriorates even more significantly than that in the indoor environment.

The performance of the SRIFO in the outdoor environment is also re-evaluated. The estimation results are demonstrated in Table \ref{tab:srifo_outdoor}. The precision is similar to that in the indoor environment. Meanwhile, the performance is also different when using different datasets, which is probably affected by the varying performance of the optical flow sensor.

Although we find slightly different results between the indoor and outdoor environments, the effectiveness of the proposed SRIBO is well proven with the aforementioned tests. This implies the SRIBO can be used for indoor and outdoor positioning with various moving ranges.

\section{Conclusion}

In this work, an efficient and resilient SRIBO framework using lightweight sensors is proposed for flying robots. Within this framework, one can perform the SRIO only using an IMU and a UWB. If an optical flow is available, an SRIFO can be further implemented for state estimation with even higher precision. The effectiveness of the proposed framework lies on two bases. First, considering aerial drag effects, the SRIBO is formulated and proven to be locally weakly observable almost surely in real-world flight, and this conclusion significantly differs from existing works. Second, the linear and dimension-reduced SRIBO is able to estimate state in a sufficiently long interval with redundant measurements, which further enhances the degree of observability as well the estimation performance. The convergence and numeric stability are also thoroughly investigated. Extensive experiments are carried out at last. It is verified the observability is effectively enhanced by considering drag effects. The SRIBO is also proven to be resilient to sensor failures and fault initialization. The overall estimation precision and computational efficiency significantly outperform the state-of-the-art approaches, and it meets the requirements of real-world applications of autonomous navigation and the relative localization of multiple agents. This framework can also be transferred to enhance the performance of the single-range based odometry for ground or underwater robots, because friction or drags also exists in those robots.

One main limitation of this work is that the covariance matrices are set empirically, which hinders the approximation of global optimization. In the future, we will try our best to improve the performance of the SRIBO regarding this aspect. In addition, external disturbances, such as wind disturbances, are also worth compensating for a further improvement of the positioning performance.

\begin{appendices}
\section{Lie Derivatives of the SRIO system}
\label{ap:lie_srio}
Regarding the dynamic model discribled with \eqref{eq:modelwithdrag}, the Lie derivative of the output is iteratvely obtained as follows
$
\small
\left\{\begin{aligned}\mathcal{L}_{f}^{0} h &=h =\frac{1}{2} \boldsymbol{p}^{\top} \boldsymbol{p} \\
\mathcal{L}_{f}^{1} h &=\nabla h \cdot \boldsymbol{f} = \boldsymbol{p}^{\top}\boldsymbol{v} \\
\mathcal{L}_{f}^{2} h &=\nabla\left[\mathcal{L}_{f}^{1} h\right] \cdot \boldsymbol{f} = \boldsymbol{v}^{T}\boldsymbol{v} + \boldsymbol{p}^{\top}(\boldsymbol{u}-\boldsymbol{\mu}\boldsymbol{v}) \\
\mathcal{L}_{f}^{3} h &=\nabla\left[\mathcal{L}_{f}^{2} h\right] \cdot \boldsymbol{f} =(3\boldsymbol{v}-\boldsymbol{\mu}\boldsymbol{p})^{\top}(\boldsymbol{u}-\boldsymbol{\mu}\boldsymbol{v})\\
\mathcal{L}_{f}^{4} h &=\nabla\left[\mathcal{L}_{f}^{3} h\right] \cdot \boldsymbol{f} =(3\boldsymbol{u}-7\boldsymbol{\mu}\boldsymbol{v}+\boldsymbol{\mu}^2\boldsymbol{p})^{\top}(\boldsymbol{u}-\boldsymbol{\mu}\boldsymbol{v}) \\
\mathcal{L}_{f}^{5} h &=\nabla\left[\mathcal{L}_{f}^{4} h\right] \cdot \boldsymbol{f}  =(15\boldsymbol{\mu}^2\boldsymbol{v}-10\boldsymbol{\mu}\boldsymbol{u}-\boldsymbol{\mu}^3\boldsymbol{p})^{\top}(\boldsymbol{u}-\boldsymbol{\mu}\boldsymbol{v}) \\
\mathcal{L}_{f}^{6} h &=\nabla\left[\mathcal{L}_{f}^{5} h\right] \cdot \boldsymbol{f}
=(25\boldsymbol{\mu}^2\boldsymbol{u}-31\boldsymbol{\mu}^3\boldsymbol{v}+\boldsymbol{\mu}^4\boldsymbol{p})^{\top}(\boldsymbol{u}-\boldsymbol{\mu}\boldsymbol{v}) \\
\end{aligned}\right.
$

\section{Lie Derivatives of the SRIFO system}
\label{ap:lie_srifo}
Regarding the dynamic model discribled with \eqref{eq:dragmodelwithflow}, the Lie derivative of the output is iteratvely obtained as follows

$
\small
\left\{\begin{array}{l}
\mathcal{L}_f^0 \boldsymbol{h}=\boldsymbol{h}=\left[\begin{array}{c} \frac{1}{2} \boldsymbol{p}^{\top} \boldsymbol{p} \\ \boldsymbol{v} \end{array}\right] \\
\mathcal{L}_f^1 \boldsymbol{h}=\nabla \boldsymbol{h} \cdot \boldsymbol{f}=\left[\begin{array}{c} \boldsymbol{p}^{\top} \boldsymbol{v} \\ \boldsymbol{u}-\boldsymbol{\mu} \boldsymbol{v} \end{array}\right] \\
\mathcal{L}_f^2 \boldsymbol{h}=\nabla\left[\mathcal{L}_f^1 \boldsymbol{h}\right] \cdot \boldsymbol{f}=\left[\begin{array}{c} \boldsymbol{v}^{\top} \boldsymbol{v}+\boldsymbol{p}^{\top}(\boldsymbol{u}-\boldsymbol{\mu} \boldsymbol{v}) \\ -\boldsymbol{\mu}(\boldsymbol{u}-\boldsymbol{\mu} \boldsymbol{v}) \end{array}\right] \\
\mathcal{L}_f^3 \boldsymbol{h}=\nabla\left[\mathcal{L}_f^2 \boldsymbol{h}\right] \cdot \boldsymbol{f}=\left[\begin{array}{c} (3 \boldsymbol{v}-\boldsymbol{\mu} \boldsymbol{p})^{\top}(\boldsymbol{u}-\boldsymbol{\mu} \boldsymbol{v}) \\ \boldsymbol{\mu}^2(\boldsymbol{u}-\boldsymbol{\mu} \boldsymbol{v}) \end{array}\right] \\
\cdots \\
\end{array}\right.
$

\section{Explicit form of $\tilde{\boldsymbol{A}}$ and $\tilde{\boldsymbol{B}}$ with Pre-integration}
\label{app:formab}
Regarding the discrete model, matrices $\tilde{\boldsymbol{A}}$ and $\tilde{\boldsymbol{B}}$ with $\ell$ steps of pre-integration are formulated as follows
$\small
\left\{\begin{array}{l}
\tilde{\boldsymbol{A}}= \begin{bmatrix}
-\boldsymbol{A}^\ell & \boldsymbol{1}_{2d} & \boldsymbol{0} & \cdots & \boldsymbol{0} \\
\boldsymbol{0} & -\boldsymbol{A}^\ell & \boldsymbol{1}_{2d} & \cdots & \boldsymbol{0} \\
\vdots & \vdots & \vdots & \vdots & \vdots \\
\boldsymbol{0} & \cdots &  \cdots & -\boldsymbol{A}^\ell & \boldsymbol{1}_{2d}  \\
\end{bmatrix}
\\
\tilde{\boldsymbol{B}}= \begin{bmatrix}
\boldsymbol{A}^{\ell-1}\boldsymbol{B} & \boldsymbol{A}^{\ell-2}\boldsymbol{B} &  \cdots & \boldsymbol{B} &\boldsymbol{0} & \cdots &\boldsymbol{0} \\
\boldsymbol{0} &\boldsymbol{A}^{\ell-1}\boldsymbol{B}  &  \cdots & \boldsymbol{A}\boldsymbol{B} & \boldsymbol{B} & \cdots &\boldsymbol{0} \\
\vdots & \vdots & \vdots & \vdots & \vdots & \vdots  \\
\boldsymbol{0} & \boldsymbol{0}  &  \cdots & \cdots & \cdots & \cdots & \boldsymbol{B} \\
\end{bmatrix}
\\
\end{array}
\right.$

\section{Parameters of the SRIBO}
\label{app:sriopar}

Parameters for the SRIBO in the indoor experiments

\vspace{0.05cm}
\begin{threeparttable}
\scriptsize
 \begin{tabular}{p{8mm}p{22mm}p{8mm}p{32mm}}
  \toprule
  \textbf{Item}   & \textbf{Value} & \textbf{Item}   & \textbf{Value} \\
  \midrule
  $\bs{\mu}$   & $\diag(1.2,2.4,4)$ & $\bs{P}^{-1}$ & $0.05\times\diag(2,1,2,2,1,2)$ \\
  $R^{-1}$  & 1 & $\bs{Q}^{-1}$ & $\diag(1,0.5,1,1,0.5,1)$ \\
  $k_t$ & 4 & $k_w$ & 38 (SRIO) / 30 (SRIFO) \\
  \bottomrule
 \end{tabular}
   \end{threeparttable}
   \vspace{0.2cm}
   
Parameters for the SRIBO in the outdoor experiments

\vspace{0.05cm}
\begin{threeparttable}
\scriptsize

 \begin{tabular}{p{8mm}p{22mm}p{8mm}p{32mm}}
  \toprule
  \textbf{Item}   & \textbf{Value} & \textbf{Item}   & \textbf{Value} \\
  \midrule
  $\bs{\mu}$   & $\diag(0.3,0.45,1.5)$ & $\bs{P}^{-1}$ & $0.05\times\diag(2,1,2,2,1,2)$ \\
  $R^{-1}$  & 1 & $\bs{Q}^{-1}$ & $\diag(1,0.5,1,1,0.5,1)$ \\
  $k_t$ & 4 & $k_w$ & 38 (SRIO) / 30 (SRIFO) \\
  \bottomrule
 \end{tabular}
\end{threeparttable}


\section{Typical measurements}
\label{app:measurecurve}



\begin{figure}[!h]
  \centering
  \subfigure[IMU]{\label{fig:imu_curve}\includegraphics[width=0.98\columnwidth]{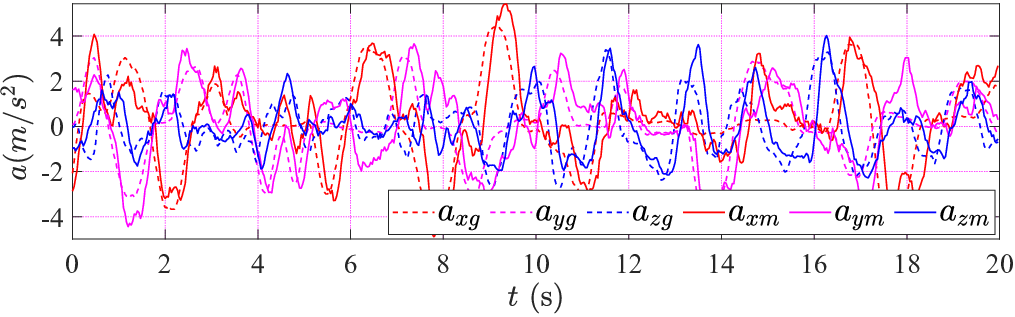}}
  \subfigure[UWB]{\label{fig:uwb_curve}\includegraphics[width=0.98\columnwidth]{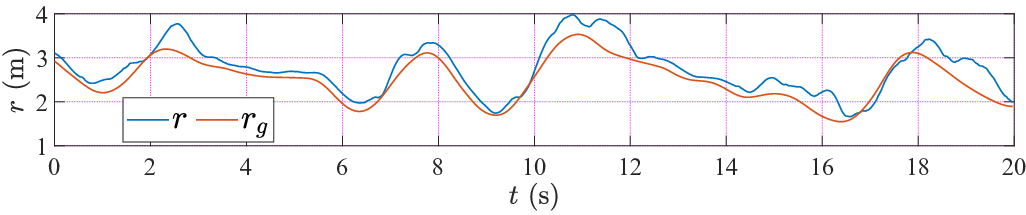}}
  \subfigure[Optical Flow ($D_0$)]{\label{fig:flow_curve.png}\includegraphics[width=0.98\columnwidth]{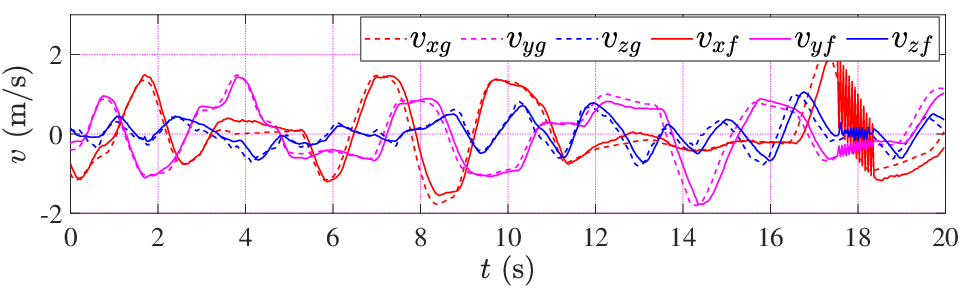}}
  \subfigure[Optical Flow ($D_2$)]{\label{fig:D2vel}\includegraphics[width=0.98\columnwidth]{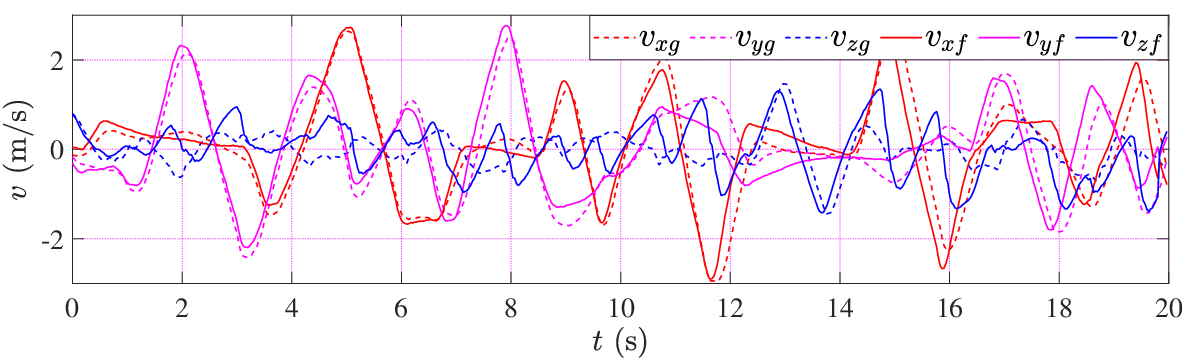}}
  \caption{Typical measurements of different sensors}\label{fig:imu_curve}
\end{figure}
\end{appendices}

\bibliographystyle{IEEEtran}
\providecommand{\noopsort}[1]{}

\end{document}